\documentclass[11pt]{article}

\usepackage[final]{acl}

\usepackage{times}
\usepackage{latexsym}
\usepackage[T1]{fontenc}
\usepackage[utf8]{inputenc}
\usepackage{microtype}
\usepackage{inconsolata}
\usepackage{enumitem}

\usepackage{graphicx}
\usepackage{amsmath}
\usepackage{amssymb}
\usepackage{amsthm}        % For formal proofs
\usepackage{booktabs}
\usepackage{multirow}
\usepackage{algorithm}     % For Algorithm box
\usepackage{algpseudocode} % For Algorithm box
\usepackage{listings}      % For Prompt display
\usepackage{xcolor}
\usepackage{subcaption}
\usepackage{makecell}      % For multiline cells in tables

\title{\textsc{EpiCaR}: Knowing What You Don’t Know Matters\\for Better Reasoning in LLMs}

\author{%
Jewon Yeom$^{1}$ \quad
Jaewon Sok$^{2}$ \quad
Seonghyeon Park$^{3}$ \quad
Jeongjae Park$^{1}$ \quad
Taesup Kim$^{1,}$\thanks{\ \ Corresponding author.} \\
$^{1}$Graduate School of Data Science, Seoul National University\\
$^{2}$Department of Rural Systems Engineering, Seoul National University\\
$^{3}$Department of Aerospace Engineering, Seoul National University
}

\begin{document}
\maketitle

\begin{abstract}
Improving the reasoning abilities of large language models (LLMs) has largely relied on iterative self-training with model-generated data. While effective at boosting accuracy, existing approaches primarily reinforce successful reasoning paths, incurring a substantial \textit{calibration cost}: models become overconfident and lose the ability to represent uncertainty. This failure has been characterized as a form of \textit{model collapse} in alignment, where predictive distributions degenerate toward low-variance point estimates.
We address this issue by reframing reasoning training as an epistemic learning problem, in which models must learn not only how to reason, but also when their reasoning should be trusted. We propose \textit{epistemically-calibrated reasoning} (\textsc{EpiCaR}) as a training objective that jointly optimizes reasoning performance and calibration, and instantiate it within an iterative supervised fine-tuning framework using explicit self-evaluation signals. Experiments on Llama-3 and Qwen-3 families demonstrate that our approach achieves Pareto-superiority over standard baselines in both accuracy and calibration, particularly in models with sufficient reasoning capacity (e.g., 3B+). This framework generalizes effectively to OOD mathematical reasoning (GSM8K) and code generation (MBPP). Ultimately, our approach enables a $3\times$ reduction in inference compute, matching the $K=30$ performance of STaR with only $K=10$ samples in capable models.
\end{abstract}

\section{Introduction}
\label{sec:intro}

\begin{figure}[t]
    \centering
    \includegraphics[width=\linewidth]{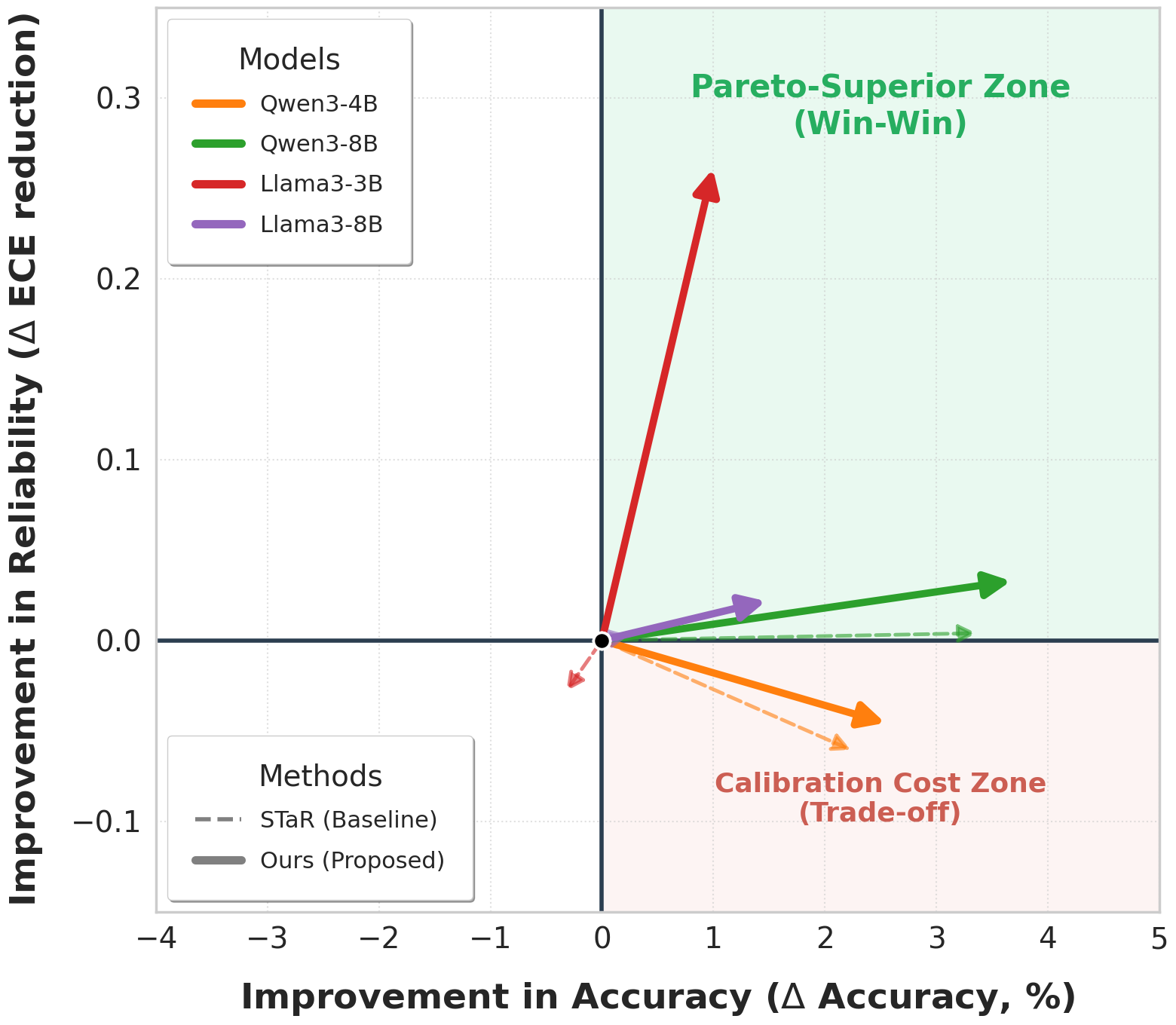} 
    \caption{\textbf{Pareto-Superior Improvement in Reasoning and Reliability.} We visualize the relative improvement in reasoning accuracy ($\Delta$ Accuracy, \%) and reliability ($\Delta$ ECE reduction) compared to the base model (at the origin). Solid and dashed arrows represent the trajectories of our proposed \textsc{EpiCaR} and the STaR baseline, respectively. While standard iterative SFT often incurs a trade-off (Calibration Cost Zone), our method consistently drives diverse model families (Llama-3 and Qwen-3) into the \textit{Pareto-Superior Zone}, achieving simultaneous gains in both task performance and uncertainty calibration.}
    \label{fig:pareto_frontier}
\end{figure}

The advent of Large Language Models (LLMs) has revolutionized complex reasoning tasks such as mathematics and logic. Techniques like Chain-of-Thought (CoT) prompting \citep{wei2022chain} have significantly boosted performance by decomposing problems into intermediate steps. However, a critical challenge remains: the discrepancy between a model's accuracy and its self-knowledge, or \textit{calibration}~\cite{kadavath2022language, guo2017calibration}. Current models frequently exhibit confident misalignment, hallucinating plausible-sounding but incorrect answers with high certainty~\cite{openai2023gpt4, lin2022teaching}. For example, a model may fail at a final arithmetic step while assigning near-certainty to its erroneous conclusion. Such behavior highlights a systematic failure to represent uncertainty about reasoning outcomes, raising critical concerns for high-stakes applications where knowing when a model should not be trusted is as important as producing a correct answer.

In the pursuit of higher reasoning accuracy, iterative self-training methods such as STaR \citep{zelikman2022star} and ReST \citep{gulcehre2023reinforced} have become a dominant paradigm, reinforcing successful reasoning paths through a ground-truth verifier. While effective for boosting raw accuracy, we argue that this positive-only feedback loop imposes a significant \textit{calibration cost} \citep{hu2025navigating} that is often overlooked. The outcome of exclusively reinforcing correct paths is a manifestation of \textit{Model Collapse} \citep{shumailov2023curse}, where the model's predictive behavior converges toward low-variance point estimates, reinforcing its own biased beliefs while discarding the distributional tails necessary for reliable uncertainty estimation.

Recent advancements in reasoning models like DeepSeek-R1 \citep{deepseekr1} have addressed reasoning performance through reinforcement learning (RL) frameworks like GRPO. Evidence suggests that such models can express their confidence more accurately by engaging in \emph{Slow Thinking} behaviors such as self-verification and backtracking within an extended CoT \citep{yoon2025reasoning}. Furthermore, \citet{zhang2025reasoning} found that models encode correctness signals within their hidden states, yet they often suffer from \emph{overthinking} and fail to exploit this information during standard generation. While effective, the reliability of these models is strictly tied to the computational overhead of increased inference-time compute.

While inference-time scaling offers a partial solution, it does not fundamentally resolve the underlying miscalibration of the base policy. In this work, we address this issue by reframing reasoning training as an epistemic learning problem, where models must learn to reason while simultaneously discerning the reliability of their own outputs. We propose \textit{epistemically-calibrated reasoning} (\textsc{EpiCaR}), a training objective that jointly optimizes reasoning performance and calibration. By instantiating \textsc{EpiCaR} within an iterative SFT framework using explicit self-evaluation signals, we enable models to navigate the trade-off between accuracy and overconfidence. As shown in Figure \ref{fig:pareto_frontier}, our framework drives models into the \textit{Pareto-Superior Zone}, matching high-sample-count performance with significantly lower inference overhead.

% ------------------------------------------------------------------
% Figure 1: Pareto Frontier Plot (Reasoning vs. Reliability)
% ------------------------------------------------------------------

% ------------------------------------------------------------------

\section{Related Work}

\paragraph{Iterative Reasoning \& Self-Improvement}
Self-training techniques like STaR \citep{zelikman2022star} and ReST \citep{gulcehre2023reinforced} bootstrap reasoning capabilities by fine-tuning on self-generated correct paths. To address the data-inefficiency of discarding incorrect attempts, V-STaR \citep{hosseini2024vstar} utilizes both correct and incorrect solutions to train a verifier that judges the correctness of generated solutions. Recent approaches explicitly target the model's meta-cognition; for instance, MASA \citep{kim2025meta} aligns the model's predictions of solution attributes (e.g., difficulty, length) with actual rollout statistics to enhance training efficiency. While V-STaR requires a separate verifier and MASA focuses on metadata for efficient gating, our method addresses the underlying \textit{calibration cost} \citep{hu2025navigating} and \emph{Model Collapse} \citep{shumailov2023curse} by internalizing the evaluation process directly into the generator's objective.

\paragraph{Calibration Cost \& Alignment Tax}
Conventional studies on the \emph{alignment tax} have almost exclusively focused on the degradation of general task performance and accuracy during the alignment process \citep{lu2024online, lin2024mitigating, fu2024disperse}. However, \citet{hu2025navigating} identify a more pervasive side effect: the \emph{calibration cost}. They argue that while the alignment tax on capability often yields mixed or inconsistent results across different benchmarks, the cost to calibration is universal, manifesting as a significant rise in overconfidence that undermines a model's reliability. Unlike the alignment tax, which concerns \textit{what} a model can do, the calibration cost represents a fundamental loss in \textit{knowing} what it knows. While \citet{hu2025navigating} propose post-hoc \textit{Model Merging} to navigate this trade-off, our approach, \textsc{EpiCaR}, mitigates this cost intrinsically during the learning process. By employing a dual-task objective that balances reasoning performance with reliability, we enable models to achieve Pareto-superiority without the need for extrinsic post-hoc interventions.

\paragraph{Calibration \& Uncertainty Estimation} 
Calibration approaches have evolved from logit scaling such as \textit{Temperature Scaling} \citep{guo2017calibration} to LLM-specific strategies like \textit{Verbalized Confidence} \citep{kadavath2022language, lin2022teaching}, though verbalized outputs remains susceptible to prompt variations \citep{xia2025influences} and struggle with long-form reasoning \citep{yang2025logu}. 
To enhance reliability, inference-time ensembles like \textit{Self-Consistency} \citep{wang2024selfconsistency} utilize response consistency but suffer from prohibitive computational costs. 
Recent literature, notably \citet{yoon2025reasoning}, identifies that reasoning models better express their confidence by leveraging \emph{Slow Thinking} behaviors such as self-verification within an extended CoT.
Furthermore, while models encode correctness in hidden states \citep{zhang2025reasoning}, and auxiliary interventions like {LitCab} \citep{liu2024litcab} or separate calibration models \citep{shen2024thermometer} have been proposed for efficiency, our approach internalizes calibration directly into the generator’s core objective. 
By balancing reasoning and self-evaluation within a unified SFT task, we achieve robust uncertainty estimation without the need for the extended inference compute.

\section{Preliminaries}
\label{sec:preliminaries}

We formally define the problem of calibrated reasoning, the mechanism for verbalized confidence estimation, and the theoretical limitations of iterative frameworks.

\subsection{Problem Formulation and Verbalized Confidence}
\label{subsec:verb_conf}
Let $\mathcal{D} = \{(x_i, y_i)\}_{i=1}^N$ be a dataset of reasoning problems $x_i$ and ground truth answers $y_i$. An LLM $P_\theta$ generates a reasoning path $r$ and a predicted answer $\hat{y}$. Our objective is to maximize $P_\theta(y | x, r)$ while ensuring a reliable confidence score $c \in [0,1]$.

Following \citet{kapoor2024large}, we adopt \textit{verbalized confidence estimation}. Instead of raw token probabilities, we prompt the model to assess its own correctness via a binary query (e.g., ``Is the answer correct? \texttt{yes/no}''). The confidence $c$ is the normalized probability of the affirmative token:
\begin{equation}
    c = \frac{P_\theta(\texttt{yes} \mid x, r, \hat{y})}{P_\theta(\texttt{yes} \mid x, r, \hat{y}) + P_\theta(\texttt{no} \mid x, r, \hat{y})}
\end{equation}
This semantic approach captures high-level certainty more effectively than sequence perplexity for reasoning tasks \citep{kuhn2023semantic}.

\subsection{Evaluation Metrics: Calibration and Discrimination}
To quantify calibration, we employ \textit{Expected Calibration Error (ECE)} \citep{guo2017calibration}, which measures the weighted average discrepancy between accuracy and confidence. Let $N$ be the total number of evaluation samples. We partition the predictions into $M$ equally-spaced intervals (bins) based on their predicted confidence scores, where $B_m$ denotes the set of indices of samples whose confidence falls into the $m$-th interval. ECE is defined as:
\begin{equation}
    \text{ECE} = \sum_{m=1}^M \frac{|B_m|}{N} \left| \text{acc}(B_m) - \text{conf}(B_m) \right|
\end{equation}
where $\text{acc}(B_m)$ and $\text{conf}(B_m)$ represent the average accuracy and the average confidence within bin $B_m$, respectively. A perfectly calibrated model achieves an $\text{ECE} = 0$.

Furthermore, we report the \textit{Brier Score} \citep{brier1950verification} as a comprehensive measure of the quality of predicted probabilities. For a set of binary outcomes (correct or incorrect), the Brier Score is defined as the mean squared error between the predicted confidence $f_i$ and the actual outcome $o_i \in \{0, 1\}$:
\begin{equation}
    \text{BS} = \frac{1}{N} \sum_{i=1}^N (f_i - o_i)^2
\end{equation}
The Brier Score is a proper scoring rule that can be decomposed into calibration, resolution, and uncertainty, providing a holistic view of the model's predictive performance. A lower Brier Score indicates a better-calibrated and more accurate model.

In addition to these metrics, we report the \textit{Area Under the Receiver Operating Characteristic curve (AUROC)} to evaluate the model's discriminative ability. While ECE measures absolute alignment, AUROC assesses the model's capacity to assign higher confidence scores to correct answers than to incorrect ones, providing a measure of reliability that is invariant to logit scaling or temperature shifts.

\subsection{The Epistemic Cost of Iterative Training}
Standard iterative methods like STaR \citep{zelikman2022star} utilize a ``positive-only'' feedback loop, fine-tuning exclusively on successful reasoning paths. We characterize the failure of this approach through the lens of the \textit{epistemic learning problem}. 

While \textit{aleatoric uncertainty} stems from inherent stochasticity, \textit{epistemic uncertainty} reflects a lack of knowledge regarding specific reasoning patterns \citep{hullermeier2021aleatoric}. By training only on correct samples, the model suffers from \emph{epistemic signal truncation}: it learns the distribution $P(r|x, y=1)$ but never encounters the decision boundary between correct and incorrect paths. This leads to a form of \textit{Model Collapse} \citep{shumailov2023curse}, where the model converges toward low-variance point estimates and discards the distributional tails necessary for representing uncertainty. Consequently, the model incurs a severe \textit{calibration cost} \citep{hu2025navigating}, manifesting as pathologically high confidence in logically flawed generations.

\section{Methodology}
\label{sec:method}

We propose \textsc{EpiCaR}, a framework that internalizes epistemic calibration into the iterative self-training loop without the need for auxiliary models or inference-time compute scaling.

\subsection{Iterative Dual-Loop Framework}
\label{sec:dual_loop}
Our framework alternates between generating reasoning traces and optimizing a unified objective that balances problem-solving and self-evaluation. The detailed procedure is outlined in Algorithm \ref{alg:dual_sft}, and training objective is presented in Section~\ref{sec:detailed_training_obj}.

\begin{algorithm}[t]
\caption{Epistemically-Calibrated Reasoning}
\label{alg:dual_sft}
\begin{algorithmic}[1]
\Require Base Model $\mathcal{M}_0$, Dataset $\mathcal{D}$, Sampling size $K$
\For{iteration $t = 1 \dots T$}
    \State $\mathcal{D}_\text{reason} \leftarrow \emptyset, \mathcal{D}_\text{eval} \leftarrow \emptyset$
    \Statex \hspace{0.5em}  \textbf{$\vartriangleright$ Step 1: Generation Phase}
    \For{$x \in \mathcal{D}$}
        \State $\hat{y}_1, \dots, \hat{y}_K \sim \text{Sample}(\mathcal{M}_{t-1}, x)$
        \For{$k = 1 \dots K$}
            \If{$\text{IsCorrect}(\hat{y}_k)$}
                \State $\mathcal{D}_\text{reason}.\text{add}(x, \hat{y}_k)$
                \State $\mathcal{D}_\text{eval}.\text{add}((x, \hat{y}_k), \texttt{yes})$
            \Else
                \State $\mathcal{D}_\text{eval}.\text{add}((x, \hat{y}_k), \texttt{no})$
            \EndIf
        \EndFor
    \EndFor
    \Statex \hspace{0.5em} \textbf{$\vartriangleright$ Step 2: Mixing Phase}
    % \Comment{1:1 Positive-Negative Ratio}
    \State $\mathcal{D}_\text{total} \leftarrow \text{Shuffle}(\mathcal{D}_\text{reason} \cup \mathcal{D}_\text{eval})$
    \Statex \hspace{0.5em}  \textbf{$\vartriangleright$Step 3: Dual-Objective Training}
    \State $\mathcal{M}_t \leftarrow \text{SFT}(\mathcal{M}_{t-1}, \mathcal{D}_\text{total})$
\EndFor
\end{algorithmic}
\end{algorithm}

\paragraph{Internalizing Evaluation Task} 
Unlike STaR, which discards incorrect generations, we utilize them as negative signals for the \textit{self-evaluation task}. For every generated path $(x, \hat{y})$, if the generation is correct, the instance is added to the \textit{reasoning task} for accuracy reinforcement and labeled as ``\texttt{yes}'' for the self-evaluation task. Conversely, if the generation is incorrect, the instance is excluded from reasoning reinforcement but labeled as ``\texttt{no}'' for the self-evaluation task.

This dual-task structure is motivated by recent findings from \citet{zhang2025reasoning}, which observed that while reasoning models encode correctness in their hidden states, standard non-reasoning models exhibit significantly degraded signals. They hypothesize that this latent capacity is acquired through exposure to reasoning patterns containing both correct and incorrect steps. By explicitly training on ``\texttt{no}'' labels for erroneous paths, our framework provides the necessary exposure to elicit and strengthen these internal signals in standard LLMs. This forces the model to exploit latent features of logical failure, effectively mitigating the alignment-induced \textit{Model Collapse} and the resulting \textit{calibration cost} described by \citet{shumailov2023curse}, and \citet{hu2025navigating}.

\subsection{Adaptive Injection Decoding (AID)}
To eliminate noise caused by formatting failures, we adapt the method from \citet{jin2025well}. When generating reasoning paths, we enforce format compliance (e.g., \verb|\boxed{}|) by injecting rigid completion strings during decoding. This ensures that valid reasoning paths are not mislabeled as ``\texttt{no}'' due to parsing errors, which would otherwise provide a confusing training signal for the self-evaluation task. We provide a stateful implementation of AID to handle edge cases such as premature termination and unclosed formatting tags; see Appendix~\ref{app:aid_details} for the detailed algorithmic logic, and Appendix~\ref{sec:ablation} for the ablation study.

\begin{table*}[t]
\centering
\small
\renewcommand{\arraystretch}{1.2}
\setlength{\tabcolsep}{5pt}
\resizebox{0.95\linewidth}{!}{%
\begin{tabular}{l l c c c c c}
\toprule
\multirow{2}{*}{\textbf{Model}} & \multirow{2}{*}{\textbf{Method}} & \textbf{Perf.} & \multicolumn{4}{c}{\textbf{Reliability \& Calibration}} \\
\cmidrule(lr){3-3} \cmidrule(lr){4-7}
& & \textbf{Acc ($\uparrow$)} & \textbf{AUROC ($\uparrow$)} & \textbf{ECE ($\downarrow$)} & \textbf{ECE \scalebox{0.5}{(+TS)} ($\downarrow$)} & \textbf{Brier ($\downarrow$)} \\
\midrule\midrule
\multirow{7}{*}{Llama-3-1B} 
& Base Model & 3.30\% & 0.525 & 0.841 & 0.516 & 0.740 \\
& \quad + Slow Thinking (ICL) & 3.00\% & 0.507 & 0.827 & 0.517 & \textbf{0.716} \\
& STaR & 3.46\% & 0.491 & 0.838 & 0.515 & 0.737 \\
& \quad + Model Merging & \textbf{3.54\%} & 0.518 & 0.838 & 0.515 & 0.737 \\
& \quad + Slow Thinking (ICL) & 3.12\% & 0.469 & \textbf{0.826} & 0.516 & \textbf{0.716} \\
& \textbf{Ours (\textsc{EpiCaR})} & 3.30\% & 0.573 & 0.871 & 0.532 & 0.800 \\
& \quad + Model Merging & 3.53\% & 0.555 & 0.848 & \textbf{0.514} & 0.754 \\
& \quad + Slow Thinking (ICL) & 3.14\% & \textbf{0.576} & 0.885 & 0.534 & 0.817 \\
\midrule
\multirow{7}{*}{Llama-3-3B} 
& Base Model & 7.56\% & 0.555 & 0.376 & 0.315 & 0.216 \\
& \quad + Slow Thinking (ICL) & 5.84\% & 0.529 & 0.605 & 0.460 & 0.424 \\
& STaR & 7.38\% & 0.562 & 0.382 & 0.344 & 0.219 \\
& \quad + Model Merging & 7.60\% & 0.562 & 0.377 & 0.318 & 0.217 \\
& \quad + Slow Thinking (ICL) & 6.24\% & 0.548 & 0.609 & 0.456 & 0.432 \\
& \textbf{Ours (\textsc{EpiCaR})} & \textbf{8.58\%} & 0.568 & \textbf{0.108} & 0.053 & \textbf{0.097} \\
& \quad + Model Merging & 7.86\% & \textbf{0.593} & 0.167 & \textbf{0.050} & 0.106 \\
& \quad + Slow Thinking (ICL) & 6.46\% & 0.571 & 0.440 & 0.436 & 0.257 \\
\midrule
\multirow{7}{*}{Llama-3-8B} 
& Base Model & 13.30\% & 0.544 & 0.496 & 0.384 & 0.368 \\
& \quad + Slow Thinking (ICL) & 12.22\% & 0.387 & 0.448 & 0.384 & 0.316 \\
& STaR & 13.46\% & 0.570 & 0.494 & 0.381 & 0.365 \\
& \quad + Model Merging & 13.72\% & 0.555 & 0.492 & 0.381 & 0.365 \\
& \quad + Slow Thinking (ICL) & 13.06\% & 0.415 & 0.436 & 0.375 & 0.312 \\
& \textbf{Ours (\textsc{EpiCaR})} & 14.42\% & \textbf{0.595} & 0.415 & 0.362 & \textbf{0.298} \\
& \quad + Model Merging & \textbf{15.02\%} & 0.571 & 0.443 & \textbf{0.361} & 0.328 \\
& \quad + Slow Thinking (ICL) & 14.38\% & 0.435 & \textbf{0.390} & \textbf{0.361} & 0.282 \\
\bottomrule
\end{tabular}
}
\caption{\textbf{Llama-3 Family Results.} \textsc{EpiCaR} generally outperforms STaR \citep{zelikman2022star} in both accuracy and calibration, with significant gains observed in the 3B and 8B variants. While the 1B model improves discriminative power (AUROC) over the baseline, it exhibits trade-offs in reasoning accuracy due to limited capacity. We compare our results with weight-space interventions \citep{hu2025navigating} and inference-time scaling \citep{yoon2025reasoning}.}
\label{tab:llama_results}
\end{table*}

\begin{table*}[t]
\centering
\small
\renewcommand{\arraystretch}{1.2}
\setlength{\tabcolsep}{6pt}
\resizebox{0.95\linewidth}{!}{%
\begin{tabular}{l l c c c c c}
\toprule
\multirow{2}{*}{\textbf{Model}} & \multirow{2}{*}{\textbf{Method}} & \textbf{Perf.} & \multicolumn{4}{c}{\textbf{Reliability \& Calibration}} \\
\cmidrule(lr){3-3} \cmidrule(lr){4-7}
& & \textbf{Acc ($\uparrow$)} & \textbf{AUROC ($\uparrow$)} & \textbf{ECE ($\downarrow$)} & \textbf{ECE \scalebox{0.5}{(+TS)} ($\downarrow$)} & \textbf{Brier ($\downarrow$)} \\
\midrule\midrule
\multirow{7}{*}{Qwen-3-1.7B} 
& Base Model & 41.44\% & 0.408 & 0.101 & 0.074 & 0.255 \\
& \quad + Slow Thinking (ICL) & 41.34\% & 0.601 & 0.091 & 0.101 & 0.245 \\
& STaR & 38.16\% & 0.430 & 0.124 & 0.102 & 0.257 \\
& \quad + Model Merging & 42.02\% & 0.413 & 0.111 & 0.072 & 0.257 \\
& \quad + Slow Thinking (ICL) & 42.98\% & 0.617 & \textbf{0.042} & 0.049 & \textbf{0.239} \\
& \textbf{Ours (\textsc{EpiCaR})} & 42.34\% & \textbf{0.637} & 0.297 & 0.079 & 0.323 \\
& \quad + Model Merging & \textbf{43.08\%} & 0.543 & 0.161 & \textbf{0.018} & 0.270 \\
& \quad + Slow Thinking (ICL) & 32.06\% & 0.631 & 0.204 & 0.054 & 0.253 \\
\midrule
\multirow{7}{*}{Qwen-3-4B} 
& Base Model & 40.66\% & 0.676 & 0.093 & 0.093 & 0.232 \\
& \quad + Slow Thinking (ICL) & 27.98\% & 0.755 & 0.275 & 0.269 & 0.240 \\
& STaR & 43.20\% & 0.765 & 0.273 & 0.154 & 0.283 \\
& \quad + Model Merging & 42.58\% & 0.784 & 0.240 & 0.168 & 0.263 \\
& \quad + Slow Thinking (ICL) & \textbf{55.78\%} & 0.796 & 0.150 & 0.131 & 0.232 \\
& \textbf{Ours (\textsc{EpiCaR})} & 43.50\% & \textbf{0.835} & 0.137 & 0.139 & 0.193 \\
& \quad + Model Merging & 43.54\% & 0.826 & 0.176 & 0.164 & 0.211 \\
& \quad + Slow Thinking (ICL) & 41.06\% & 0.820 & \textbf{0.126} & \textbf{0.080} & \textbf{0.186} \\
\midrule
\multirow{7}{*}{Qwen-3-8B} 
& Base Model & 45.86\% & 0.727 & 0.196 & 0.121 & 0.259 \\
& \quad + Slow Thinking (ICL) & 55.16\% & 0.673 & 0.134 & 0.065 & 0.251 \\
& STaR & 49.52\% & 0.710 & 0.179 & 0.117 & 0.258 \\
& \quad + Model Merging & 48.08\% & 0.712 & 0.190 & 0.122 & 0.261 \\
& \quad + Slow Thinking (ICL) & 54.52\% & 0.659 & 0.190 & \textbf{0.032} & 0.271 \\
& \textbf{Ours (\textsc{EpiCaR})} & 49.76\% & \textbf{0.797} & 0.131 & 0.088 & \textbf{0.206} \\
& \quad + Model Merging & 49.72\% & 0.769 & 0.123 & 0.103 & 0.217 \\
& \quad + Slow Thinking (ICL) & \textbf{55.56\%} & 0.780 & \textbf{0.107} & 0.148 & 0.220 \\
\bottomrule
\end{tabular}
}
\caption{\textbf{Qwen-3 Family Results.} \textsc{EpiCaR} demonstrates superior discriminative reliability (AUROC) across all scales and outperforms STaR \citep{zelikman2022star}, especially at 3B and 8B scale. Smaller models exhibit mixed results in calibration error and scaling efficiency compared to the baseline. We compare our results with weight-space interventions \citep{hu2025navigating} and inference-time scaling \citep{yoon2025reasoning}.}
\label{tab:qwen_results}
\end{table*}

\section{Experimental Setup}
\label{sec:experimental_setup}

\paragraph{Datasets} We evaluate our framework using the \textbf{MATH}\footnote{\url{https://huggingface.co/datasets/EleutherAI/hendrycks_math}} dataset \citep{hendrycks2021measuring} for the main iterative training loop ($T=3$). We test out-of-distribution (OOD) generalization on \textbf{GSM8K}\footnote{\url{https://huggingface.co/datasets/openai/gsm8k}} \citep{cobbe2021training} and cross-domain robustness on \textbf{MBPP}\footnote{\url{https://huggingface.co/datasets/google-research-datasets/mbpp}} \citep{austin2021program}. For calibration tuning, we use a 500-instance split-validation for MATH, and use the official validation set for MBPP. Specifically, for the scaling analysis in Section~\ref{sec:inference-time-ensemble}, we evaluate on the \textbf{MATH-500}\footnote{\url{https://huggingface.co/datasets/HuggingFaceH4/MATH-500}} subset to ensure comparability with prior literature. For MBPP, we adopt a 3-shot prompting setup and evaluate functional correctness via a sandboxed execution environment; see Appendix~\ref{app:mbpp_details} for details on code extraction and the verification pipeline.

\paragraph{Models and Baselines} We employ the \textbf{Llama-3}~\citep{dubey2024llama3} and \textbf{Qwen-3}~\citep{yang2025qwen3} families, focusing on 8B variants for MBPP and ablation studies. We compare our method against three primary baselines: (1) the \textbf{Base Model}, (2) \textbf{STaR} \citep{zelikman2022star} for iterative self-improvement, and (3) \textbf{Slow Thinking (ICL)} \citep{yoon2025reasoning} for reliability. Additionally, we evaluate post-hoc \textbf{Model Merging} \citep{hu2025navigating} by sweeping $\lambda \in \{0.0, 0.2, \dots, 1.0\}$ to navigate the alignment-calibration frontier. Detailed algorithmic backgrounds and implementation for these baselines are provided in Appendix~\ref{app:baselines}.

\paragraph{Inference-time Scaling Protocol} To evaluate how internalized calibration scales with inference-time compute, we sample $K \in \{1, 10, 30\}$ reasoning paths. We compare two aggregation strategies: (1) \textbf{Self-Consistency (SC)} \citep{wang2022self}, which selects the plurality answer via simple majority voting, and (2) \textbf{Confidence-Informed Self-Consistency (CISC)} \citep{taubenfeld-etal-2025-confidence}. Unlike SC, CISC performs a weighted majority vote by utilizing the model's self-assessed confidence $\mathcal{V}(r_i)$ as a fidelity signal. This allows the model to surpass the accuracy of high-sample-count SC with significantly fewer samples (e.g., $K=10$ matching $K=30$), thereby reducing inference compute. The formal definition of the CISC aggregation and normalization process is provided in Appendix~\ref{app:cisc_details}.

\paragraph{Implementation} Performance is measured via {Accuracy}, {AUROC}, {ECE}, and {Brier Score}. To assess the model's intrinsic uncertainty beyond logit-level shifts, we also report ECE after applying \emph{Temperature Scaling (TS)} \citep{guo2017calibration}. Training uses LoRA ($r=16, \alpha=32$) on 4x NVIDIA H100 GPUs. Following \citet{kapoor2024large}, we update hidden features to support accurate verbalized confidence. Detailed hyperparameters, prompting templates, and the optimization procedure for TS are provided in Appendix \ref{appendix:full_setup}, and Appendix~\ref{appendix:ts_details}.

\section{Results and Analysis}

In this section, we evaluate our method's performance across different model families and sizes. We report Accuracy for reasoning capability, and AUROC, ECE, and Brier Score for calibration quality. To ensure a rigorous comparison, we contrast our \textsc{EpiCaR} against the standard {STaR} baseline, as well as state-of-the-art inference-time scaling (\textit{Slow Thinking} \citep{yoon2025reasoning}) and weight-space intervention (\textit{Model Merging} \citep{hu2025navigating}) strategies. For \emph{Model Merging}, we report the results obtained using the optimal $\lambda$ that yields the best result with respect to accuracy and calibration, following~\citet{hu2025navigating}. See Appendix~\ref{appendix:model_merging_full} for full results for all $\lambda$.

\subsection{Main Findings: Accuracy and Reliability Synergy}
Our results across both model families (Tables \ref{tab:llama_results} and \ref{tab:qwen_results}) demonstrate that internalizing the self-evaluation objective allows LLMs to navigate the reasoning-reliability frontier more effectively, particularly as model scale increases.

\paragraph{Mitigating Calibration Cost of Iterative SFT}
Standard iterative SFT ({STaR}) consistently incurs a calibration cost, improving accuracy while often degrading discriminative reliability. For instance, in Llama-3-1B, while STaR marginally boosts accuracy, it drops AUROC to 0.491. In contrast, our approach recovers this discriminative power (AUROC 0.573), albeit with a slight trade-off in reasoning accuracy at this scale. However, for larger models like Llama-3-3B and 8B, our method achieves strict Pareto-dominance, effectively preventing the overconfident model collapse typical of positive-only feedback loops (e.g., reducing ECE from 0.376 to 0.108 in Llama-3-3B). Notably, \textsc{EpiCaR} achieves the lowest Brier Score in most cases (e.g., 0.097 for Llama-3-3B), indicating superior overall predictive quality.

\paragraph{Foundation for Inference-time Scaling}
A pivotal finding is the synergy between our internalized calibration and \textit{Slow Thinking} \citep{yoon2025reasoning} in capable models. While Slow Thinking generally boosts performance, it can be unstable on uncalibrated models. Our calibrated framework stabilizes this behavior in the 8B variants, achieving a performance of 55.56\% on Qwen-3-8B. We note, however, that this synergy is scale-dependent; in intermediate sizes (e.g., Qwen-3-4B), the added inference complexity does not yield performance gains over STaR, suggesting that a critical mass of reasoning capability is required to effectively leverage the self-evaluation signal.

\paragraph{Weight-space Interpolation \& Merging}
We observe significant synergy with \textit{Model Merging} \citep{hu2025navigating}, indicating that our reliability signals are complementary to weight-space interventions. While STaR yields marginal gains from merging, applying it to our checkpoints unlocks substantial headroom: the \emph{Ours + Merging} variant achieves the highest Llama-3-8B accuracy of 15.02\%. Precision calibration is also enhanced, with Qwen-3-1.7B reaching a near-perfect ECE (+TS) of 0.018, demonstrating the flexibility of our method for post-hoc optimization.

\begin{table}[t]
\centering
\small
\renewcommand{\arraystretch}{1.1}
\setlength{\tabcolsep}{5pt}
\resizebox{\columnwidth}{!}{
\begin{tabular}{l l c c c c}
\toprule
\textbf{Family} & \textbf{Method} & \textbf{Acc. ($\uparrow$)} & \textbf{AUROC ($\uparrow$)} & \textbf{ECE ($\downarrow$)} & \textbf{Brier ($\downarrow$)} \\
\midrule\midrule
\multirow{3}{*}{Llama-3-1B}   
& Base Model & 3.49\% & 0.545 & 0.841 & 0.741 \\
& STaR & 3.49\% & 0.478 & \textbf{0.840} & \textbf{0.740} \\
& \textbf{Ours (\textsc{EpiCaR})} & \textbf{3.64\%} & \textbf{0.645} & 0.922 & 0.887 \\
\midrule
\multirow{3}{*}{Llama-3-3B}   
& Base Model & 14.94\% & 0.527 & 0.303 & 0.223 \\
& STaR & 17.13\% & 0.497 & 0.284 & 0.228 \\
& \textbf{Ours (\textsc{EpiCaR})} & \textbf{21.46\%} & \textbf{0.606} & \textbf{0.020} & \textbf{0.166} \\
\midrule
\multirow{3}{*}{Llama-3-8B}   
& Base Model & 31.39\% & \textbf{0.565} & 0.329 & 0.324 \\
& STaR & 32.15\% & 0.561 & 0.338 & 0.332 \\
& \textbf{Ours (\textsc{EpiCaR})} & \textbf{37.45\%} & \textbf{0.565} & \textbf{0.223} & \textbf{0.284} \\
\midrule
\midrule
\multirow{3}{*}{Qwen-3-1.7B}   
& Base Model & 77.33\% & 0.490 & \textbf{0.445} & \textbf{0.376} \\
& STaR & 78.54\% & 0.493 & 0.541 & 0.463 \\
& \textbf{Ours (\textsc{EpiCaR})} & \textbf{79.30\%} & \textbf{0.660} & 0.671 & 0.610 \\
\midrule
\multirow{3}{*}{Qwen-3-4B}   
& Base Model & 85.06\% & 0.611 & 0.364 & 0.255 \\
& STaR & 86.05\% & 0.693 & \textbf{0.099} & 0.122 \\
& \textbf{Ours (\textsc{EpiCaR})} & \textbf{88.25\%} & \textbf{0.756} & 0.132 & \textbf{0.106} \\
\midrule
\multirow{3}{*}{Qwen-3-8B}   
& Base Model & 85.90\% & 0.654 & \textbf{0.215} & 0.162 \\
& STaR & 87.95\% & 0.678 & \textbf{0.215} & \textbf{0.147} \\
& \textbf{Ours (\textsc{EpiCaR})} & \textbf{89.46\%} & \textbf{0.722} & 0.364 & 0.216 \\
\bottomrule
\end{tabular}
} 
\caption{\textbf{Zero-shot GSM8K Results.} \textsc{EpiCaR} consistently improves reasoning accuracy while maintaining or enhancing AUROC.}
\label{tab:gsm8k_with_brier}
\end{table}

\subsection{Out-of-Distribution Generalization}

To evaluate the generalization performance of our internalized calibration objective, we conduct zero-shot evaluations on GSM8K \citep{cobbe2021training}. This serves as a critical out-of-distribution (OOD) benchmark, as the models were trained exclusively on the MATH dataset. 

\paragraph{Reliability of Verbalized Confidence Rankings}
As shown in Table \ref{tab:gsm8k_with_brier}, our method consistently enhances the model's discriminative reliability across both families. A key finding is that while standard iterative self-training ({STaR}) often leads to a degradation in AUROC (e.g., Llama-3-1B falling to 0.478), our approach significantly improves it, reaching 0.645 and 0.606 for Llama-3-1B and 3B, respectively. This demonstrates that the verbalized confidence scores produced by our framework are better separated, assigning substantially higher probabilities of correctness to true reasoning paths than to erroneous ones, even in unseen domains.

\paragraph{Discriminative Power vs. Absolute Calibration}
Despite high AUROC, some variants exhibit elevated {Raw ECE}. We hypothesize that while our dual-objective training effectively teaches the model to \textit{distinguish} between correct and incorrect paths, it does not explicitly supervise the absolute alignment of verbalized probability logits with empirical accuracy. Consequently, the model maintains a strong sense of \textit{relative} certainty, but the absolute values remain somewhat detached from the true probability space. We identify the explicit optimization of these verbalized probability logits as a crucial direction for future work to bridge the gap between discriminative power and absolute calibration.

\subsection{Generalization to Other Tasks}
Finally, we examine whether our approach generalizes beyond mathematics to the domain of code generation. We apply our framework to the MBPP benchmark \citep{austin2021program}, focusing on the 8B model variants.

\begin{table}[h]
\centering
\small
\renewcommand{\arraystretch}{1.1}
\setlength{\tabcolsep}{5pt}
\resizebox{\columnwidth}{!}{
\begin{tabular}{l l c c c c}
\toprule
\textbf{Model} & \textbf{Method} & \textbf{Acc. ($\uparrow$)} & \textbf{AUROC ($\uparrow$)} & \textbf{ECE \scalebox{0.5}{(+TS)} ($\downarrow$)} & \textbf{Brier ($\downarrow$)} \\
\midrule\midrule
\multirow{3}{*}{Llama-3-8B} 
& Base Model & 37.35\% & \textbf{0.551} & 0.398 & 0.391 \\
& STaR & 37.74\% & 0.523 & 0.390 & 0.387 \\
& \textbf{Ours (\textsc{EpiCaR})} & \textbf{39.30\%} & 0.538 & \textbf{0.113} & \textbf{0.246} \\
\midrule
\multirow{3}{*}{Qwen-3-8B} 
& Base Model & 45.14\% & 0.622 & 0.066 & \textbf{0.286} \\
& STaR & 45.14\% & 0.577 & 0.066 & 0.285 \\
& \textbf{Ours (\textsc{EpiCaR})} & \textbf{45.91\%} & \textbf{0.628} & \textbf{0.059} & 0.285 \\
\bottomrule
\end{tabular}
}
\caption{\textbf{Results on MBPP coding task.} Our method (\textsc{EpiCaR}) consistently improves Pass Rate (Accuracy) and demonstrates competitive calibration performance across different model families.}
\label{tab:mbpp_results_with_brier}
\end{table}

Table \ref{tab:mbpp_results_with_brier} summarizes the performance across three stages: {Base Model}, {STaR}, and {Ours}. While our method does not always achieve the absolute highest AUROC, it consistently outperforms the STaR baseline in discriminative reliability. Specifically, in Llama-3-8B, STaR incurs a significant calibration cost, dropping AUROC from 0.551 to 0.523, whereas our method mitigates this degradation, recovering it to 0.538. 

Crucially, \textsc{EpiCaR} achieves a significantly lower Brier Score (0.246 vs. 0.387 in Llama-3-8B) and the lowest ECE (+TS) across both families, reaching 0.113 for Llama-3-8B and 0.059 for Qwen-3-8B. This improvement in a proper scoring rule (Brier Score) confirms that internalizing self-evaluation signals effectively prevents the overconfident model collapse typical of positive-only self-training, even in cross-domain reasoning tasks like programming.

\subsection{Inference-time Scaling Performance}
\label{sec:inference-time-ensemble}

We further investigate how internalized calibration impacts the inference-time scaling laws of reasoning models. By ensembling $K$ sampled reasoning paths, we analyze the synergy between internalized confidence and test-time compute.

\begin{figure}[t]
\centering
\includegraphics[width=\columnwidth]{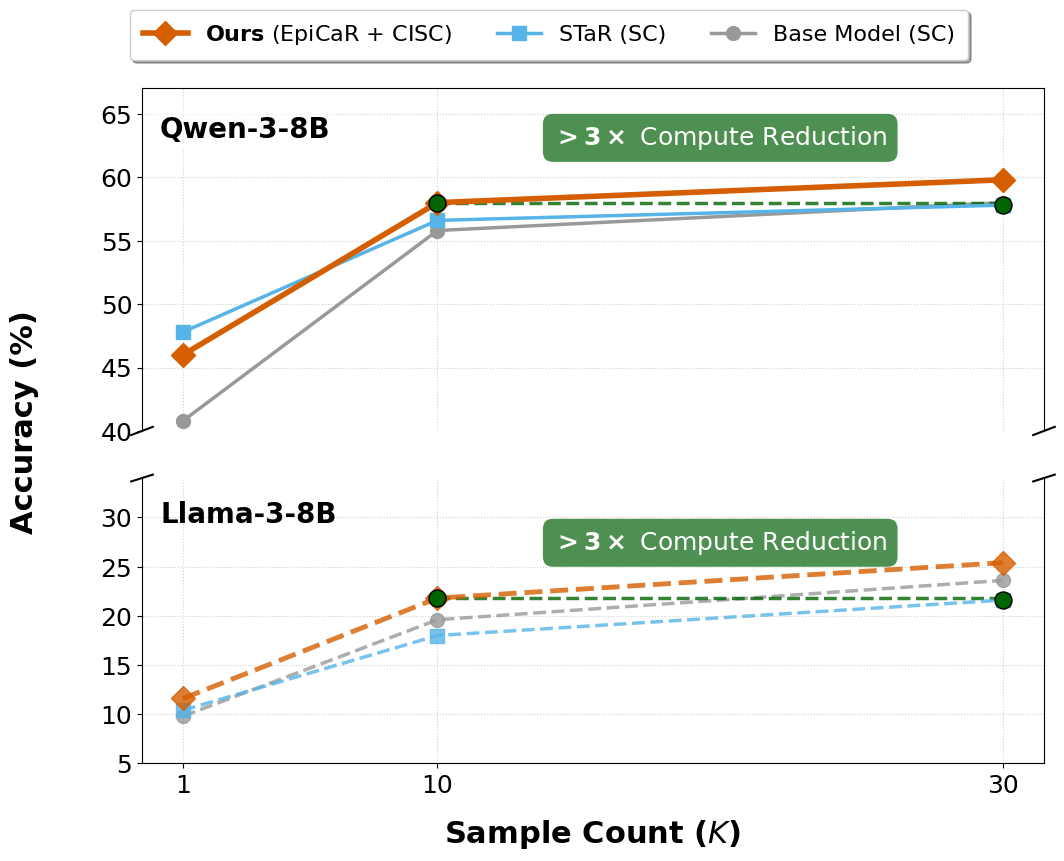}
\caption{\textbf{Inference-time Scaling on MATH-500.} Visual representation of ensemble accuracy across sample sizes $K$. Our framework paired with CISC achieves superior scaling efficiency, outperforming STaR and establishing a new frontier for compute-optimal reasoning.}
\label{fig:main_accuracy_scaling}
\end{figure}

Figure~\ref{fig:main_accuracy_scaling} illustrates the results on the MATH-500 benchmark. Our framework demonstrates superior scaling efficiency compared to the STaR baseline. Notably, when paired with CISC \cite{taubenfeld-etal-2025-confidence}, our 8B model achieves a $3\times$ reduction in inference compute—matching or exceeding the $K=30$ performance of STaR with only $K=10$ samples. This suggests that \textsc{EpiCaR} provides a more robust foundation for compute scaling, as internalized reliability signals effectively suppress erroneous consensus paths that often cause performance saturation in uncalibrated models. A comprehensive analysis of all scales and reliability trajectories is provided in Appendix~\ref{appendix:full_inference_ensembling}.

\section{Conclusion}

In this work, we addressed the \textit{calibration cost} inherent in iterative self-training, where models gain reasoning accuracy at the expense of reliability and uncertainty representation. We introduced \textsc{EpiCaR}, a framework that reframes reasoning training as an epistemic learning task by internalizing a dual-objective of accuracy reinforcement and explicit self-evaluation. Our experiments across Llama-3 and Qwen-3 families demonstrate that \textsc{EpiCaR} consistently achieves \textit{Pareto-superiority}, simultaneously enhancing both performance and calibration while generalizing robustly to OOD math (GSM8K) and code generation (MBPP). Furthermore, we established that internalized calibration serves as a critical driver for inference-time scaling efficiency; by leveraging self-assessed confidence through weighted ensembling (CISC), our models match high-sample-count performance with significantly fewer reasoning paths, yielding an effective $3\times$ reduction in inference compute. These findings suggest that teaching models to ``know what they don't know'' is not merely a post-hoc safety constraint but a fundamental prerequisite for building compute-optimal reasoning systems. Ultimately, this work advocates for a paradigm shift in model alignment, where uncertainty calibration is treated as an integral training objective for trustworthy reasoning agents.

\newpage
\section*{Limitations}

Despite the performance gains and improved calibration, our work has several limitations. 

\paragraph{Domain Scope and Verification.} While we demonstrated that \textsc{EpiCaR} generalizes across mathematics (MATH, GSM8K) and code generation (MBPP), our evaluation remains centered on domains with objective, automated ground-truth verification. It remains to be seen whether \textsc{EpiCaR} can be effectively applied to more subjective or ambiguous domains, such as legal reasoning or creative writing, where defining a clear signal for knowing what it does not know is inherently more challenging.

\paragraph{Model Scale and Capacity Constraints.} Our experiments were conducted on models up to 8B parameters. As identified in Appendix \ref{appendix:full_inference_ensembling}, our method is sensitive to the baseline reasoning capacity. In extremely low-accuracy regimes, such as Llama-3-1B, the scarcity of successful reasoning paths during the iterative loop can prevent the model from learning the nuanced discriminative features necessary for accurate self-evaluation. This suggests a critical mass threshold for effective internalized calibration, and the dynamics may further differ in ultra-large scale models or specialized frontier models.

\paragraph{Generalization Gap in Absolute Calibration.} We observe a notable discrepancy between discriminative power and absolute numerical calibration in out-of-distribution (OOD) scenarios. While \textsc{EpiCaR} consistently maintains superior AUROC across benchmarks like GSM8K—indicating that it generalizes the relative ranking of reasoning paths effectively—the absolute probabilities it verbalizes (measured by Raw ECE) exhibit higher volatility compared to in-distribution results on MATH. This suggests that while the model successfully semanticizes its internal certainty to distinguish correctness, the precise mapping from these semantic tokens to empirical probabilities is sensitive to domain shifts. This highlights a fundamental gap between learning to rank outputs and achieving absolute logit alignment in unseen domains.

\paragraph{Comparison with RL and Verifier-based Methods.} While recent RL-based approaches like RLCR \citep{damani2025rlcr} achieve calibration through specialized reward structures, they often suffer from training instability and high hyperparameter sensitivity. In contrast, \textsc{EpiCaR} operates within a stable iterative SFT framework. Furthermore, unlike V-STaR \citep{hosseini2024vstar}, which requires a separate verifier model to filter reasoning paths, our approach unifies generation and self-evaluation into a single model. While this integration is the key driver behind our reported $3\times$ reduction in inference compute by eliminating the overhead of multiple model passes, it may be bounded by the internal capacity of the single model compared to systems with dedicated external verification components.

\paragraph{Potential for Reinforcement Learning Integration.} A promising direction for future research is the integration of our internalized calibration objective into reinforcement learning (RL) frameworks. The self-evaluation signals generated in our dual-loop framework could serve as an intrinsic reward mechanism or a dense feedback signal for RL algorithms such as PPO~\cite{schulman2017proximal}, DPO~\cite{rafailov2023direct}, or GRPO~\cite{deepseekr1}. By using the model's own epistemic certainty to weight or mask rewards during exploration, it may be possible to further stabilize the learning of complex reasoning trajectories and mitigate the overoptimization issues common in standard RLHF pipelines.

\paragraph{Sensitivity of the Verbalization Paradigm.} \textsc{EpiCaR} adopts the verbalized confidence paradigm, following evidence that LLMs can semanticize internal certainty more effectively than logit-based indicators \citep{kadavath2022language, lin2022teaching}. While this approach captures high-level semantic certainty, it remains susceptible to the well-documented phenomenon of prompt sensitivity \citep{xia2025influences}. Although our results demonstrate that \textsc{EpiCaR} significantly improves the calibration of these verbalized signals, further research is required to ensure their absolute robustness across a wider range of elicitation templates and linguistic contexts to prevent self-evaluation from being biased by phrasing.

\section*{Ethical Considerations}

Our research aims to improve LLM reliability, a critical step toward safe AI deployment. However, we acknowledge several ethical implications. 

\paragraph{Misuse and Over-reliance:} By enhancing a model's ability to express confidence, there is a risk that users may over-rely on outputs when the model reports high certainty. While \textsc{EpiCaR} significantly reduces overconfidence, no technique is infallible, and "confident hallucinations" may still occur, potentially leading to errors in high-stakes decision-making.

\paragraph{Bias Amplification:} As our framework utilizes iterative self-training on model-generated data, there is a potential to amplify biases present in the pre-trained state or the training distribution. We recommend applying our method alongside rigorous bias-detection and mitigation protocols. 

\paragraph{Environmental Impact:} The iterative nature of our framework involves multiple rounds of generation and fine-tuning, incurring higher computational costs during training compared to single-pass SFT. However, we mitigate this impact by demonstrating that our models achieve superior reliability with significantly reduced inference-time compute via efficient scaling.

\paragraph{Use of AI Assistants.}
In accordance with the ACL Policy on AI Assistance, we acknowledge the use of Gemini 3 Pro\footnote{\url{https://deepmind.google/technologies/gemini/}} to assist with code debugging and writing polishing. All experimental designs, data analyses, and scientific claims presented in this work were verified by the authors.

\bibliography{custom}
\clearpage
\appendix

\section{Full Results and Analysis of Inference-time Ensembling}
\label{appendix:full_inference_ensembling}

\subsection{Comprehensive Ensemble Accuracy of All Scales}

We perform an extensive evaluation of the scaling properties of \textit{EpiCaR} across all model variants using the \textbf{MATH-500} test set. By varying the number of sampled reasoning paths $K \in \{1, 10, 30\}$, we map the compute-performance frontier as summarized in Table~\ref{tab:comprehensive_results_final}.

\paragraph{Internalized Calibration as a Scaling Foundation.} 
Our experimental results consistently demonstrate that our method (\textbf{Ours}) establishes a superior and more stable baseline for inference-time scaling compared to both the pre-trained Base Model and the STaR baseline. At the 8B parameter scale, the synergy between internalized calibration and weighted aggregation becomes most apparent. Pairing our model with \textbf{CISC} \cite{taubenfeld-etal-2025-confidence} achieves state-of-the-art results: 25.40\% for Llama-3 and 59.80\% for Qwen-3 at $K=30$. This trajectory suggests that when a model is trained to generate its own reliability signals, the performance gains from additional inference compute do not just accumulate—they accelerate, as the model's self-evaluation serves as a high-fidelity filter for the sampling process.

\paragraph{Synergy, Error Suppression, and Saturation.}
The synergy between EpiCaR and the CISC aggregation protocol is a pivotal finding. While standard Self-Consistency (SC) \cite{wang2022self} relies on a frequentist plurality vote, it is vulnerable to "common errors" where incorrect reasoning paths form a deceptive majority. As observed in the Qwen-3 8B results, the STaR baseline exhibits significant performance saturation, reaching a plateau at 57.80\% for $K=30$. In contrast, our calibrated model continues to scale effectively, effectively breaking the saturation ceiling. This confirms that a model that "knows what it knows" can utilize verbalized confidence to suppress high-frequency erroneous paths. By assigning low weights to incorrect but common reasoning traces, EpiCaR ensures that the ensemble consensus is driven by logical fidelity rather than mere repetition.

\paragraph{The Critical Mass Threshold in Reasoning capacity.}
A nuanced and critical finding is that the advantages of internalized calibration are not uniform across all model scales, identifying what we term a \textit{"Critical Mass Threshold."} In the extremely low-accuracy regime, such as with \textbf{Llama-3 1B} (baseline accuracy $\sim$1.80\%), the gap between our method and the STaR baseline diminishes. We hypothesize that when the model's initial reasoning capability is too sparse, the training process lacks a sufficient density of correct reasoning paths to learn the subtle discriminative features required for accurate self-evaluation. This suggests that while EpiCaR is a powerful scaling tool, its benefits are most salient once a model reaches a fundamental level of reasoning competence. Future work should investigate whether specialized data augmentation can lower this threshold for smaller model architectures.

\begin{table*}[h]
\centering
\small
\renewcommand{\arraystretch}{1.1}
\resizebox{0.8\textwidth}{!}{
\begin{tabular}{l l l l c c c}
\toprule
\textbf{Family} & \textbf{Size} & \textbf{Method} & \textbf{Ensemble} & \boldmath{$K=1$} & \boldmath{$K=10$} & \boldmath{$K=30$} \\
\midrule \midrule
\multirow{18}{*}{Llama-3} & \multirow{6}{*}{1B} & \multirow{2}{*}{Base Model} & SC & \multirow{2}{*}{1.80\%} & 2.00\% & 3.80\% \\
& & & CISC & & 2.40\% & \textbf{4.40\%} \\
\cmidrule{3-7}
& & \multirow{2}{*}{STaR} & SC & \multirow{2}{*}{\textbf{3.20\%}} & \textbf{3.20\%} & \textbf{4.40\%} \\
& & & CISC & & 3.00\% & 3.80\% \\
\cmidrule{3-7}
& & \multirow{2}{*}{Ours} & SC & \multirow{2}{*}{1.60\%} & 2.80\% & 3.40\% \\
& & & CISC & & \textbf{3.20\%} & 3.40\% \\
\cmidrule{2-7}
& \multirow{6}{*}{3B} & \multirow{2}{*}{Base Model} & SC & \multirow{2}{*}{5.40\%} & 8.20\% & 10.60\% \\
& & & CISC & & 8.40\% & 10.20\% \\
\cmidrule{3-7}
& & \multirow{2}{*}{STaR} & SC & \multirow{2}{*}{4.80\%} & 8.40\% & \textbf{15.00\%} \\
& & & CISC & & 9.40\% & 14.60\% \\
\cmidrule{3-7}
& & \multirow{2}{*}{Ours} & SC & \multirow{2}{*}{\textbf{7.40\%}} & \textbf{13.60\%} & 13.40\% \\
& & & CISC & & \textbf{13.60\%} & 13.20\% \\
\cmidrule{2-7}
& \multirow{6}{*}{8B} & \multirow{2}{*}{Base Model} & SC & \multirow{2}{*}{9.80\%} & 19.60\% & 23.60\% \\
& & & CISC & & 20.20\% & 23.40\% \\
\cmidrule{3-7}
& & \multirow{2}{*}{STaR} & SC & \multirow{2}{*}{10.40\%} & 18.00\% & 21.60\% \\
& & & CISC & & 17.80\% & 21.80\% \\
\cmidrule{3-7}
& & \multirow{2}{*}{Ours} & SC & \multirow{2}{*}{\textbf{11.60\%}} & 21.00\% & 24.80\% \\
& & & CISC & & \textbf{21.80\%} & \textbf{25.40\%} \\
\midrule \midrule
\multirow{18}{*}{Qwen-3} & \multirow{6}{*}{1.7B} & \multirow{2}{*}{Base Model} & SC & \multirow{2}{*}{29.60\%} & 33.40\% & 39.60\% \\
& & & CISC & & 35.80\% & 39.80\% \\
\cmidrule{3-7}
& & \multirow{2}{*}{STaR} & SC & \multirow{2}{*}{37.20\%} & 46.60\% & 48.80\% \\
& & & CISC & & 46.80\% & 47.40\% \\
\cmidrule{3-7}
& & \multirow{2}{*}{Ours} & SC & \multirow{2}{*}{\textbf{40.40\%}} & \textbf{51.40\%} & \textbf{53.00\%} \\
& & & CISC & & 51.20\% & 52.40\% \\
\cmidrule{2-7}
& \multirow{6}{*}{4B} & \multirow{2}{*}{Base Model} & SC & \multirow{2}{*}{36.60\%} & 51.40\% & 54.60\% \\
& & & CISC & & 51.60\% & 55.40\% \\
\cmidrule{3-7}
& & \multirow{2}{*}{STaR} & SC & \multirow{2}{*}{41.80\%} & 53.00\% & 54.60\% \\
& & & CISC & & 53.60\% & 55.00\% \\
\cmidrule{3-7}
& & \multirow{2}{*}{Ours} & SC & \multirow{2}{*}{\textbf{44.40\%}} & 54.20\% & 53.80\% \\
& & & CISC & & \textbf{55.00\%} & \textbf{57.20\%} \\
\cmidrule{2-7}
& \multirow{6}{*}{8B} & \multirow{2}{*}{Base Model} & SC & \multirow{2}{*}{40.80\%} & 55.80\% & 58.00\% \\
& & & CISC & & 55.80\% & 58.40\% \\
\cmidrule{3-7}
& & \multirow{2}{*}{STaR} & SC & \multirow{2}{*}{\textbf{47.80\%}} & 56.60\% & 57.80\% \\
& & & CISC & & 56.80\% & 58.00\% \\
\cmidrule{3-7}
& & \multirow{2}{*}{Ours} & SC & \multirow{2}{*}{46.00\%} & 57.40\% & 59.20\% \\
& & & CISC & & \textbf{58.00\%} & \textbf{59.80\%} \\
\bottomrule
\end{tabular}
}
\caption{\textbf{Comprehensive Ensemble Accuracy on MATH-500 (\%).} Comparison of unweighted Self-Consistency (SC \cite{wang2022self}) and Confidence-Informed Self-Consistency (CISC \cite{taubenfeld-etal-2025-confidence}) across multiple model families and sizes. $K=1$ values are merged as the methods are identical without multiple samples. Bolding indicates the highest accuracy within each size group for a specific $K$.}
\label{tab:comprehensive_results_final}
\end{table*}

\subsection{Full Reliability Analysis of Scaling: SC vs. CISC}
\label{appendix:reliability_sc_cisc}

To assess the qualitative aspects of uncertainty estimation within an ensemble, we evaluate reliability metrics across $K \in \{5, 10, 30\}$. For the \textbf{CISC} variants, we derive a scalar ensemble confidence score $\mathcal{C}_{CISC}$ for each unique answer $a$:
\begin{equation}
\mathcal{C}_{CISC}(a) = \frac{\sum_{i: \text{ans}(r_i) = a} \mathcal{V}(r_i)}{\sum_{j=1}^K \mathcal{V}(r_j)}
\end{equation}
where $\mathcal{V}(r_i) \in [0, 1]$ represents the model's internalized verbalized confidence for path $r_i$. The comprehensive results for AUROC and ECE are detailed in Table~\ref{tab:reliability_scaling_full_comparison}.

\paragraph{Reliability Collapse in STaR Baseline.} 
A striking phenomenon observed in our scaling analysis is the \textbf{Reliability Collapse} of the STaR baseline as $K$ increases. For instance, in Llama-3 8B, STaR's AUROC degrades significantly from 0.7895 ($K=5$) to 0.7387 ($K=30$). This degradation suggests a fundamental flaw in positive-only iterative training: by exclusively reinforcing successful paths, the model homogenizes its error patterns, leading to incorrect reasoning paths that form a deceptively high-confidence consensus. As more samples are added, these "confident hallucinations" dominate the ensemble, undermining its discriminative power. In sharp contrast, our model (\textbf{Ours}) sustains or even improves its AUROC as compute scales. By internalizing logical failure modes during training, EpiCaR maintains a clear margin between correct and incorrect reasoning paths, which is essential for the trustworthy deployment of LLMs in high-stakes reasoning tasks.

\paragraph{Metric Divergence: The Interplay of ECE and AUROC.}
We observe a characteristic divergence in reliability trends as $K$ scales: while ECE generally improves, AUROC tends to decay. We hypothesize that this is driven by two counteracting statistical forces:
\begin{itemize}
    \item \textbf{Statistical Smoothing:} ECE benefits from larger sample sizes as the averaged ensemble confidence naturally converges toward the true population accuracy, reducing absolute calibration error.
    \item \textbf{Consensus-Driven Noise:} AUROC suffers as the probability of incorrect paths forming a high-confidence plurality increases with $K$, narrowing the discriminative gap between true and false positives.
\end{itemize}
Our framework effectively mitigates the decay rate of AUROC compared to all baselines. By ensuring that the model's confidence is tied to the \textit{intrinsic logic} of the path rather than the \textit{extrinsic frequency} of the answer, EpiCaR achieves a more robust reliability-compute trade-off.

\begin{table*}[h]
\centering
\small
\renewcommand{\arraystretch}{1.1}
\resizebox{\textwidth}{!}{
\begin{tabular}{l l l l | c c | c c | c c}
\toprule
\multirow{2}{*}{\textbf{Family}} & \multirow{2}{*}{\textbf{Size}} & \multirow{2}{*}{\textbf{Method}} & \multirow{2}{*}{\textbf{Ensemble}} & \multicolumn{2}{c|}{\boldmath{$K=5$}} & \multicolumn{2}{c|}{\boldmath{$K=10$}} & \multicolumn{2}{c}{\boldmath{$K=30$}} \\
& & & & AUROC $\uparrow$ & ECE $\downarrow$ & AUROC $\uparrow$ & ECE $\downarrow$ & AUROC $\uparrow$ & ECE $\downarrow$ \\
\midrule \midrule
\multirow{18}{*}{Llama-3} & \multirow{6}{*}{1B} 
& \multirow{2}{*}{Base} & SC & 0.4784 & 0.2380 & 0.4760 & 0.1644 & 0.6490 & 0.0995 \\
& & & CISC & 0.4824 & 0.2416 & 0.5442 & 0.1633 & 0.6213 & 0.0941 \\
\cmidrule{3-10}
& & \multirow{2}{*}{STaR} & SC & \textbf{0.6870} & \textbf{0.2172} & 0.7390 & \textbf{0.1490} & 0.6643 & \textbf{0.0915} \\
& & & CISC & 0.6848 & 0.2247 & \textbf{0.7540} & 0.1539 & \textbf{0.7308} & 0.0981 \\
\cmidrule{3-10}
& & \multirow{2}{*}{\textbf{Ours}} & SC & 0.5636 & 0.2516 & 0.6398 & 0.1742 & 0.6478 & 0.1273 \\
& & & CISC & 0.5564 & 0.2551 & 0.7078 & 0.1731 & 0.6588 & 0.1284 \\
\cmidrule{2-10}
& \multirow{6}{*}{3B} 
& \multirow{2}{*}{Base} & SC & 0.7002 & 0.2252 & 0.7225 & 0.1450 & 0.6760 & 0.0843 \\
& & & CISC & 0.7758 & 0.2488 & 0.7390 & 0.1544 & 0.6843 & 0.0875 \\
\cmidrule{3-10}
& & \multirow{2}{*}{STaR} & SC & 0.7413 & 0.2356 & 0.6972 & 0.1470 & 0.6169 & \textbf{0.0460} \\
& & & CISC & 0.6967 & 0.2540 & 0.6942 & 0.1458 & 0.6343 & 0.0558 \\
\cmidrule{3-10}
& & \multirow{2}{*}{\textbf{Ours}} & SC & \textbf{0.8170} & \textbf{0.2080} & \textbf{0.7556} & \textbf{0.1212} & 0.7561 & 0.0951 \\
& & & CISC & 0.7475 & 0.2618 & 0.7303 & 0.1468 & \textbf{0.7972} & 0.0913 \\
\cmidrule{2-10}
& \multirow{6}{*}{8B} 
& \multirow{2}{*}{Base} & SC & 0.7470 & 0.1952 & 0.7283 & 0.0978 & 0.7184 & 0.0472 \\
& & & CISC & 0.7499 & 0.2108 & 0.7034 & 0.1083 & 0.7059 & 0.0613 \\
\cmidrule{3-10}
& & \multirow{2}{*}{STaR} & SC & 0.7895 & 0.1812 & 0.7741 & 0.1058 & 0.7387 & 0.0529 \\
& & & CISC & 0.7892 & 0.2057 & 0.7810 & 0.1164 & 0.7300 & 0.0452 \\
\cmidrule{3-10}
& & \multirow{2}{*}{\textbf{Ours}} & SC & 0.7824 & \textbf{0.1628} & \textbf{0.7988} & \textbf{0.0684} & \textbf{0.7722} & 0.0397 \\
& & & CISC & \textbf{0.8511} & 0.2175 & 0.7795 & 0.0782 & 0.7636 & \textbf{0.0383} \\
\midrule \midrule
\multirow{18}{*}{Qwen-3} & \multirow{6}{*}{1.7B} 
& \multirow{2}{*}{Base} & SC & 0.7315 & 0.2156 & 0.7182 & 0.1508 & 0.7065 & 0.1181 \\
& & & CISC & 0.7414 & 0.2303 & 0.7208 & 0.1624 & 0.6967 & 0.1057 \\
\cmidrule{3-10}
& & \multirow{2}{*}{STaR} & SC & 0.7638 & 0.1960 & 0.7657 & 0.1344 & 0.7521 & 0.1174 \\
& & & CISC & 0.7771 & 0.2177 & 0.7572 & 0.1307 & 0.7688 & 0.1157 \\
\cmidrule{3-10}
& & \multirow{2}{*}{\textbf{Ours}} & SC & 0.7853 & \textbf{0.1672} & 0.7756 & \textbf{0.0986} & 0.7750 & \textbf{0.0900} \\
& & & CISC & \textbf{0.7950} & 0.1860 & \textbf{0.7774} & 0.1186 & \textbf{0.7812} & 0.0930 \\
\cmidrule{2-10}
& \multirow{6}{*}{4B} 
& \multirow{2}{*}{Base} & SC & 0.7869 & \textbf{0.1304} & 0.7830 & \textbf{0.0968} & 0.7667 & \textbf{0.0573} \\
& & & CISC & 0.7834 & 0.1360 & \textbf{0.7954} & 0.1015 & 0.7721 & 0.0478 \\
\cmidrule{3-10}
& & \multirow{2}{*}{STaR} & SC & 0.8170 & 0.2028 & 0.7864 & 0.1390 & 0.7713 & 0.1229 \\
& & & CISC & \textbf{0.8266} & 0.2063 & 0.7916 & 0.1431 & 0.7762 & 0.1295 \\
\cmidrule{3-10}
& & \multirow{2}{*}{\textbf{Ours}} & SC & 0.7737 & 0.1776 & 0.7544 & 0.1242 & \textbf{0.7859} & 0.1108 \\
& & & CISC & 0.7764 & 0.1915 & 0.7776 & 0.1503 & 0.7781 & 0.1139 \\
\cmidrule{2-10}
& \multirow{6}{*}{8B} 
& \multirow{2}{*}{Base} & SC & 0.7626 & \textbf{0.0884} & 0.7213 & \textbf{0.0786} & 0.7457 & \textbf{0.0767} \\
& & & CISC & 0.7791 & 0.1205 & 0.7359 & 0.0945 & 0.7516 & 0.0707 \\
\cmidrule{3-10}
& & \multirow{2}{*}{STaR} & SC & 0.7972 & 0.1632 & 0.7735 & 0.1260 & 0.7854 & 0.1009 \\
& & & CISC & \textbf{0.8128} & 0.1598 & 0.7919 & 0.1296 & 0.7909 & 0.1018 \\
\cmidrule{3-10}
& & \multirow{2}{*}{\textbf{Ours}} & SC & 0.7767 & 0.1640 & 0.7926 & 0.1098 & 0.7914 & 0.0959 \\
& & & CISC & 0.7932 & 0.1630 & \textbf{0.8077} & 0.1201 & \textbf{0.8010} & 0.0998 \\
\bottomrule
\end{tabular}
}
\caption{\textbf{Comprehensive Reliability Scaling across All Models (AUROC and ECE).} Comparison of frequency-based (SC) and weighted (CISC) confidence estimates across training stages. Bold entries indicate the best performance (highest AUROC or lowest ECE) within each size group for a given $K$.}
\label{tab:reliability_scaling_full_comparison}
\end{table*}

\clearpage

\section{Full Experimental Configurations}
\label{appendix:full_setup}

\subsection{Iterative Training Loop Details}
For both the STaR baseline and EpiCaR, we conduct three complete iterations ($T=3$) of the self-improvement loop. 
In each iteration, the model generates $K$ candidate reasoning paths for each problem in the MATH training set. For STaR, only paths reaching the ground-truth answer are retained for fine-tuning. For our method, we perform the mixing strategy (Section ~\ref{sec:dual_loop}) using both correct and incorrect attempts to refine the self-evaluation objective. We find that performance and calibration metrics typically stabilize after the third iteration, after which further training often leads to marginal returns.

\subsection{Data Processing and Validation Protocol}
For the iterative SFT loop, we utilize the training split of the \textbf{MATH} dataset (12,500 problems). To evaluate the robustness of internalized calibration in out-of-distribution (OOD) scenarios, we test on \textbf{GSM8K} without further fine-tuning. 
A critical component of our evaluation is the split-validation protocol. Calibration metrics can be overly optimistic if the scaling parameters are optimized on the same set used for reporting. Thus, we randomly sample $N=500$ instances from the test set as a "calibration-validation" set. We find the optimal temperature $T$ that minimizes ECE on this set and apply it to the remaining test instances ($N_{test}-500$) for final reporting.

\subsection{Data Licensing and Usage}
We utilize standard public benchmarks consistent with their intended research purposes. The \textbf{MATH} \citep{hendrycks2021measuring} and \textbf{GSM8K} \citep{cobbe2021training} datasets are released under the MIT License, while the \textbf{MBPP} \citep{austin2021program} dataset is distributed under the CC-BY 4.0 License.

\subsection{Baseline Implementation Details}
\begin{itemize}[leftmargin=1.5em]
    \item \textbf{STaR:} We implement the standard self-taught reasoner by filtering self-generated reasoning paths that reach the ground-truth answer. The model is fine-tuned for 3 iterations, with the checkpoint from $T_{i}$ serving as the generator for $T_{i+1}$.
    \item \textbf{Slow Thinking (ICL):} We adopt the few-shot prompts from \citet{yoon2025reasoning}, which encourage models to "think slow" by verbalizing internal verification steps. This baseline represents the upper bound of inference-time calibration without training.
    \item \textbf{Model Merging:} We use weight-space interpolation between the base model ($\theta_{base}$) and SFT checkpoints ($\theta_{SFT}$). The coefficient $\lambda$ represents the weight of the SFT model.
\end{itemize}

\subsection{Training and Hyperparameters}
All models were fine-tuned using the Hugging Face TRL library. We employ Parameter-Efficient Fine-Tuning (PEFT) via LoRA to ensure scalability.
\begin{table}[h]
\centering
\small
\begin{tabular}{ll}
\toprule
\textbf{Hyperparameter} & \textbf{Value} \\
\midrule\midrule
Optimizer & AdamW \\
Learning Rate & $1 \times 10^{-5}$ \\
LR Scheduler & Cosine \\
Batch Size & 32 \\
Max Sequence Length & 2048 \\
LoRA Rank ($r$) & 16 \\
LoRA Alpha ($\alpha$) & 32 \\
LoRA Target Modules & All linear layers \\
Precision & bfloat16 \\
\bottomrule
\end{tabular}
\caption{Detailed Hyperparameters for Dual-Objective Iterative SFT.}
\end{table}

\subsection{Inference Implementation}
To evaluate the effectiveness of our proposed method in terms of both reasoning accuracy and confidence calibration, we implemented a high-throughput inference pipeline using the \texttt{vLLM} library \citep{kwon2023efficient}. All models were loaded in \texttt{bfloat16} precision with eager execution mode enabled to ensure stability. 

\subsection{Hyperparameters and Decoding Strategies}
We utilized distinct decoding strategies tailored to the nature of each benchmark:

\begin{itemize}
    \item \textbf{Greedy Decoding for Standard Benchmarks (MATH, GSM8K):} For the primary evaluation of reasoning accuracy and calibration, we employed greedy decoding (\texttt{temperature=0.0}). The generation max length was set to 1024 tokens to allow sufficient reasoning steps (CoT). To accurately extract answers and measure verbalized confidence, we utilized a prefix injection technique (e.g., ``\textit{So, the answer is} $\backslash$boxed\{'') with a forced stop token at the closing brace. This injection step acts as a constraint that improves the format compliance and accuracy of the final answer.
    
    \item \textbf{Sampling for Robustness (MATH-500):} To assess the model's performance under self-consistency \citep{wang2022self}, we employed sampling with \texttt{temperature=0.7} and generated $N=30$ candidate paths per problem. The \texttt{repetition\_penalty} was dynamically adjusted to maintain diversity without degradation.
    
    \item \textbf{Code Generation (MBPP):} For the code generation tasks, we adopted a few-shot prompting strategy consistent with prior work \citep{austin2021program}. The inference was performed with \texttt{temperature=0.0} for pass@1 evaluation, employing specific stop sequences (e.g., ``Problem:'', ``Tests:'') to prevent hallucination beyond the solution function.
\end{itemize}

\section{Implementation Details of Adaptive Injection Decoding (AID)}
\label{app:aid_details}

Our \textit{Adaptive Injection Processor} is implemented as a stateful \texttt{LogitsProcessor} that monitors the decoding process at each step. Unlike static constraints, it dynamically transitions through states based on the model's output. The operational logic is categorized into three primary mechanisms:

\paragraph{1. Triggering and Injection State}
The processor maintains an \texttt{is\_injecting} flag and an \texttt{injection\_step} counter. The injection is triggered under two conditions: (1) when the model attempts to emit a termination token (EOS) before a \verb|\boxed{}| tag is generated, or (2) when the sequence reaches a soft length limit (e.g., $L_{max} - 150$ tokens). During the injection state, the model's logits are overridden to force the generation of the transition phrase: \textit{"\textbackslash nSo, the answer is \textbackslash boxed\{"}.

\paragraph{2. Stateful Format Enforcement}
To ensure the injected format is correctly finalized, the processor tracks the sequence for the existence of the \texttt{boxed} token and its corresponding closing brace (\verb|}|). 
\begin{itemize}
    \item \textbf{Premature EOS Prevention:} If the model attempts to terminate while a box is open (\texttt{has\_injected} is True but no closing brace is found), the processor overrides the EOS token with a forced closing brace (\verb|}|).
    \item \textbf{Content Length Control:} To prevent infinite or malformed generation within the answer box, the processor enforces a maximum content length ($C_{max}=40$). If exceeded, it mandates a closing brace and subsequent termination.
\end{itemize}

\paragraph{3. Zombie and Sampling Defense}
The processor utilizes a \texttt{finished\_mask} to prevent "zombie" generations where a model might continue after a logical EOS. Additionally, in cases where no box is detected, the processor suppresses all stop tokens until the injection logic is triggered, ensuring that every sample produced for the training pipeline is parsable and evaluatable.

\paragraph{AID as a De-noising Filter.}
Our ablation study (Section~\ref{sec:ablation}) shows a performance collapse without Adaptive Injection Decoding (AID). We clarify that AID does not introduce reasoning bias; rather, it acts as a sanitization process to remove label noise. Without AID, a mathematically sound derivation might be labeled as ``\texttt{no}'' due to minor formatting errors (e.g., missing a closing \verb|\boxed{}|). Such mislabeling provides a catastrophic signal, teaching the model that ``valid logic is incorrect.'' AID ensures that the ``\texttt{no}'' label strictly reflects logical failures, thereby protecting the integrity of the self-evaluation supervision.

\section{Details of MBPP Evaluation and Execution}
\label{app:mbpp_details}

For the code generation task, we implement a robust evaluation pipeline to ensure that the model's reasoning capabilities are not underestimated due to minor formatting inconsistencies.

\paragraph{1. Robust Code Extraction}
Language models often produce conversational fillers or incomplete code snippets. As shown in our implementation, we employ a multi-stage extraction heuristic:
\begin{itemize}
    \item \textbf{Block Extraction:} We first attempt to parse the output for Markdown-style Python blocks (\verb|```python ... ```|).
    \item \textbf{Keyword-based Recovery:} If no block is found, the processor searches for functional keywords such as \texttt{import}, \texttt{def}, or \texttt{return} to identify the start of the logic.
    \item \textbf{Signature Injection:} A common failure mode in LLMs is omitting the function header. To mitigate this, our pipeline checks if the canonical function signature (e.g., \texttt{def remove\_Occ(...)}) is present; if missing, it automatically prepends the signature to the model's output to ensure the script is functionally complete and ready for execution.
\end{itemize}

\paragraph{2. Sandboxed Execution and Verification}
Correctness is verified by executing the extracted code against a set of hidden unit tests provided by the MBPP dataset.
\begin{itemize}
    \item \textbf{Environment:} Execution is performed in a separate subprocess using Python's \texttt{multiprocessing} to prevent side effects and handle potential infinite loops via a 3-second timeout.
    \item \textbf{Dependency Injection:} To support diverse coding problems, the environment is pre-loaded with essential libraries, including \texttt{math}, \texttt{heapq}, \texttt{collections}, and \texttt{itertools}.
    \item \textbf{Strict Pass Criterion:} A task is labeled as \textit{Correct} ($y=1$) if and only if the code executes without errors and passes every \texttt{assert} statement in the test list.
\end{itemize}

\paragraph{3. Prompting Strategy}
We use a 3-shot prompt consisting of diverse Python tasks (e.g., sum of squares, average of cubes) to steer the model toward a consistent "Reasoning-then-Solution" format. This mirrors the iterative self-training objective used in our math reasoning tasks, allowing for a direct comparison of calibration performance across different domains.

\section{Technical Overview of Baselines and Related Concepts}
\label{app:baselines}

In this section, we provide detailed technical backgrounds for the primary baselines and concepts utilized in our study: STaR, Slow Thinking (ICL), and Model Merging.

\subsection{Self-Taught Reasoner (STaR)}
\textbf{STaR} \citep{zelikman2022star} is an iterative bootstrapping framework designed to improve a model's reasoning capabilities using only a large dataset of questions and answers without intermediate rationales. Its core loop consists of:
\begin{enumerate}
    \item \textbf{Rationale Generation:} The model is prompted (via few-shot) to generate a Chain-of-Thought (CoT) and a final answer.
    \item \textbf{Filtering:} Only rationales that lead to the correct ground-truth answer are retained.
    \item \textbf{Rationalization:} For questions where the model fails, it is provided with the correct answer as a "hint" to generate a backward-reasoned rationale.
    \item \textbf{Fine-tuning:} The model is fine-tuned on the combined set of successful and rationalized paths.
\end{enumerate}
In our experiments, STaR serves as the primary iterative baseline, representing the "accuracy-first" reinforcement approach that often incurs a high calibration cost.

\subsection{Slow Thinking via In-Context Learning (ICL)}
Based on the findings of \citet{yoon2025reasoning}, \textbf{Slow Thinking} refers to the cognitive behaviors exhibited by advanced reasoning models (e.g., DeepSeek-R1, o1), such as exploring alternative approaches, self-verification, and backtracking within an extended CoT.
\begin{itemize}
    \item \textbf{Mechanism:} Unlike standard CoT, slow thinking is non-linear. The model dynamically adjusts its internal confidence as it rejects erroneous paths or confirms logical steps.
    \item \textbf{ICL Implementation:} For non-reasoning models (baselines), we elicit these behaviors via In-Context Learning (ICL) by providing few-shot exemplars that demonstrate these "slow" cognitive patterns. This serves as a strong baseline for reliability without parameter updates.
\end{itemize}

\paragraph{Prompt Templates.}
\label{appendix:prompts}

For the \textit{Slow Thinking (ICL)} baseline, we adopt the prompting strategy from \citet{yoon2025reasoning}\footnote{\url{https://github.com/MattYoon/reasoning-models-confidence}}. This template encourages the model to engage in meta-cognitive reasoning within \texttt{<think>} tags before providing a final answer. The full template used for our experiments is provided in Listing \ref{lst:slow_thinking_prompt}.

% language={}를 사용하여 plain text임을 명시합니다.
\begin{lstlisting}[language={}, caption={Full Few-Shot Prompt Template for Slow Thinking (ICL)}, label={lst:slow_thinking_prompt}]
First, generate your thought process aloud in first-person in between the <think> tags. Then, outside of the <think> tag, generate your final answer. You must stop thinking outside of the <think> tag, and commit to providing an answer in the format of **Answer**: $ANSWER. Here are some examples.

[Example 1]
Question: Who was the next British Prime Minister after Arthur Balfour?
Output:
<think>
Okay, so I need to figure out who was the next British Prime Minister after Arthur Balfour... [Detailed reasoning process] ...So, the sequence is Balfour, then Campbell-Bannerman, then Asquith.
</think>
**Answer**: Henry Campbell-Bannerman
[Example 1 End]

[Example 2]
Question: Men Against the Sea and Pitcairn's Island were two sequels to what famous novel?
Output:
<think>
... [Detailed reasoning process] ...So the famous novel they're sequels to is "Mutiny on the Bounty."
</think>
**Answer**: Mutiny on the Bounty.
[Example 2 End]

[Example 3]
Question: River Phoenix died during the making of which movie?
Output:
<think>
... [Detailed reasoning process] ...So, to sum up my thoughts: Paramount was originally called the "Hartford Film Company."
</think>
**Answer**: Hartford Film Company.
[Example 3 End]

This is the end of the examples. The following is the question you need to answer following the style of the examples.
\end{lstlisting}

\subsection{Alignment-Calibration Trade-off and Model Merging}
\citet{hu2025navigating} formalize the \textbf{Alignment Tax} not just as a loss of task performance, but as a \textbf{Calibration Cost}—where models become universally overconfident after instruction tuning.
\begin{itemize}
    \item \textbf{Model Merging:} To mitigate this, they propose interpolating the weights of a Pre-Trained (PT) base model ($\theta_{PT}$) and its Instruction-Tuned (IT) counterpart ($\theta_{IT}$):
    \begin{equation}
        \theta_{merged} = (1 - \lambda)\theta_{PT} + \lambda\theta_{IT}
    \end{equation}
    where $\lambda$ is the merging coefficient. 
    \item \textbf{Pareto-Superior Frontier:} This simple linear interpolation often reveals a "sweet spot" where the merged model achieves higher accuracy than both parents while substantially recovering the calibration lost during alignment. We use this post-hoc method to benchmark the limits of weight-space optimization against our intrinsic training-time approach, \textit{EpiCaR}.
\end{itemize}

\section{Formal Definition of Confidence-Informed Self-Consistency (CISC)}
\label{app:cisc_details}

To maximize the utility of our model's internalized calibration during inference, we adopt \textit{Confidence-Informed Self-Consistency (CISC)} \citep{taubenfeld-etal-2025-confidence} as our primary scaling strategy. Given a question $q$ and a set of $m$ sampled reasoning paths and answers $\{(r_1, a_1), \dots, (r_m, a_m)\}$, the CISC aggregation process consists of three stages:

\paragraph{1. Confidence Extraction.} For each reasoning path $r_i$, we extract the verbalized confidence score $c_i = \mathcal{V}(r_i)$. In our framework, these scores are generated via the verbalized confidence described in Section~\ref{subsec:verb_conf}.

\paragraph{2. Softmax Normalization.} To balance the relative importance of answer frequency (majority) and individual path reliability, the raw confidence scores are normalized using a temperature-scaled softmax:
\begin{equation}
    \tilde{c}_i = \frac{\exp(c_i / T)}{\sum_{j=1}^m \exp(c_j / T)}
\end{equation}
where $T$ is a tunable hyper-parameter. As $T \to \infty$, CISC collapses to vanilla SC (uniform weights). As $T \to 0$, the mechanism prioritizes the single path with the highest confidence, effectively becoming a "greedy" confidence selection.

\paragraph{3. Weighted Aggregation.} The final answer $\hat{a}_{CISC}$ is determined by summing the normalized confidence weights for each unique answer candidate:
\begin{equation}
    \hat{a}_{CISC} = \arg \max_a \sum_{i=1}^m \mathbb{1}[a_i = a] \cdot \tilde{c}_i
\end{equation}

\section{Calibration and Temperature Scaling Details}
\label{appendix:ts_details}

In addition to standard calibration metrics, we employ \textit{Temperature Scaling (TS)} \citep{guo2017calibration} to evaluate the quality of the model's confidence scores independently of its overconfidence bias. 

\paragraph{Temperature Scaling Procedure.} 
We optimize a single scalar parameter $T > 0$ for each model family and dataset. The temperature $T$ is determined by minimizing the Negative Log-Likelihood (NLL) on a held-out validation set ($N_{val}=500$):
\begin{equation}
    \min_{T} -\sum_{i=1}^{N_{val}} \log \left( \sigma(\mathbf{z}_i / T) \right)
\end{equation}
where $\mathbf{z}_i$ represents the model's output logits for the correct answer. We denote the ECE calculated using these calibrated probabilities as \textbf{ECE-TS}. This allows us to distinguish between models that are inherently uncalibrated and those that simply require a linear adjustment to their confidence variance.

\section{Extended Related Work}

\paragraph{Comparison with RL and Verifier-based Methods.}
While recent RL-based approaches like RLCR \citep{damani2025rlcr} achieve calibration, they often suffer from training instability and high hyperparameter sensitivity. In contrast, \textsc{EpiCaR} operates within a stable iterative SFT framework. Furthermore, unlike V-STaR \citep{hosseini2024vstar}, which requires a separate verifier model, our approach unifies generation and verification into a single model. This integration is the key driver behind our reported $3\times$ reduction in inference compute, as it eliminates the overhead of managing multiple model forward passes.

\paragraph{Prompting vs. Fine-tuning for Calibration.}
Recent studies have debated whether LLMs can express uncertainty through zero-shot prompting alone. While some suggest that scale improves self-evaluation \citep{kadavath2022language}, \citet{kapoor2024large} argue that prompting on its own is insufficient for reliable calibration due to high sensitivity to linguistic variances. They demonstrate that fine-tuning on even a small graded dataset significantly outperforms black-box prompting by leveraging the model's internal features. \textsc{EpiCaR} builds on this insight by integrating this calibration-tuning into an iterative reasoning loop, ensuring that the model's confidence is grounded in its evolving reasoning capacity rather than static prompt templates.

\section{Detailed Training Objective}
\label{sec:detailed_training_obj}
We prioritize a minimalist design to ensure robustness. Unlike multi-task frameworks that require tuned auxiliary coefficients, \textsc{EpiCaR} utilizes the standard Causal Language Modeling (CLM) loss:
\begin{equation}
    \mathcal{L}_{\text{EpiCaR}} = \mathcal{L}_{CLM} = -\sum_{t} \log P_\theta(w_t \mid w_{<t})
\end{equation}
The model treats reasoning tokens and the self-evaluation token (e.g., \texttt{yes/no}) as a single continuous sequence.

\paragraph{Intrinsic Curriculum via Natural Mixing.}
Rather than using a fixed mixing weight, we adopt a natural mixing strategy. By pooling all reasoning and self-evaluation samples as described in Algorithm 1, the training distribution dynamically evolves. In early iterations where model accuracy is low, the dataset is naturally dominated by negative self-evaluation samples (``\texttt{no}''). This forces the model to prioritize self-objectivity and uncertainty representation (epistemic grounding). As reasoning capability improves, the proportion of positive reasoning samples naturally increases. This serves as an intrinsic curriculum that stabilizes the learning trajectory without manual intervention.

\section{Ablation Study: Impact of Adaptive Injection Decoding}
\label{sec:ablation}
To validate the structural importance of our proposed \textit{Adaptive Injection Decoding (AID)}, we perform an ablation analysis using Llama-3-8B and Qwen-3-8B on the MATH dataset. We compare our full method (\textbf{Ours}) against a variant trained without injection decoding (\textbf{w/o AID}), where formatting errors are treated as standard incorrect samples.

\begin{table}[h]
\centering
\small
\renewcommand{\arraystretch}{1.1}
\setlength{\tabcolsep}{6pt}
\resizebox{\columnwidth}{!}{
\begin{tabular}{l l c c c}
\toprule
\textbf{Model} & \textbf{Variant} & \textbf{Acc. ($\uparrow$)} & \textbf{AUROC ($\uparrow$)} & \textbf{ECE ($\downarrow$)} \\
\midrule\midrule
\multirow{2}{*}{Llama-3-8B} & \textbf{Ours (Full)} & \textbf{14.47\%} & \textbf{0.594} & \textbf{0.415} \\
 & w/o AID & 2.56\% & 0.507 & 0.580 \\
\midrule
\multirow{2}{*}{Qwen-3-8B} & \textbf{Ours (Full)} & \textbf{49.62\%} & \textbf{0.800} & \textbf{0.133} \\
 & w/o AID & 47.83\% & 0.784 & 0.152 \\
\bottomrule
\end{tabular}
}
\caption{\textbf{Ablation on Adaptive Injection Decoding (AID).} Removing AID leads to significant performance degradation, highlighting its role in preventing formatting noise from corrupting the evaluation signal.}
\label{tab:ablation}
\end{table}

As shown in Table \ref{tab:ablation}, the removal of AID results in a significant degradation in both reasoning accuracy and calibration metrics. 
For Llama-3-8B, the accuracy drops catastrophically to 2.56\%. This collapse occurs because, without AID, valid reasoning paths that suffer from minor formatting issues (e.g., missing box delimiters) are mislabeled as "incorrect" (``\texttt{no}''). This introduces severe label noise, confusing the model's self-evaluation mechanism and destabilizing the reinforcement loop. AID effectively filters this noise, ensuring that the negative signal purely reflects \textit{logical} failures rather than syntactic ones.

\section{Full Results for Model Merging}
\label{appendix:model_merging_full}

In this section, we provide the comprehensive results of our model merging experiments across the Llama-3 and Qwen-3 model families. Table~\ref{tab:llama_merging_full} and Table~\ref{tab:qwen_merging_full} details the performance and reliability metrics for varying interpolation coefficients $\lambda$, ranging from the pure base model ($\lambda=0.0$) to the fully fine-tuned models ($\lambda=1.0$).

\paragraph{Effect of Interpolation Coefficient ($\lambda$)} 
The results demonstrate that model merging—interpolating weights between the base model and the fine-tuned model—serves as an effective regularizer. For most model scales, we observe that an intermediate value of $\lambda$ (e.g., $0.6 \leq \lambda \leq 0.8$) often provides a superior balance between task-specific accuracy and probabilistic calibration. Interestingly, in larger models such as Qwen-3-8B and Llama-3.1-8B, our method (\textsc{EpiCaR}) maintains high discriminative performance (AUROC) even at $\lambda=1.0$, whereas STaR often exhibits signs of overconfidence or degraded calibration as the fine-tuning progresses.

\paragraph{Comparison: STaR vs. \textsc{EpiCaR}} 
Across all tested architectures, \textsc{EpiCaR} consistently outperforms the STaR baseline in both discriminative reliability (AUROC) and calibration error (ECE and Brier Score). Notably:
\begin{itemize}
    \item \textbf{Discriminative Power:} \textsc{EpiCaR} achieves significantly higher AUROC scores, particularly at the 4B and 8B scales. This indicates that the epistemic uncertainty captured during our training process allows the model to better distinguish between correct and incorrect reasoning paths.
    \item \textbf{Calibration Stability:} While STaR often shows fluctuating ECE values as $\lambda$ increases, \textsc{EpiCaR} tends to show a more stable or improving trend in Brier scores. This suggests that the calibration objective in \textsc{EpiCaR} successfully aligns the model's confidence with its actual predictive correctness.
\end{itemize}

\paragraph{Scaling Trends} 
We observe a clear scaling trend where larger models (8B) benefit more from the weight-space intervention. While smaller models like Llama-3.2-1B show mixed results in calibration error, they still exhibit improved AUROC when merged with our epistemically-calibrated weights. This confirms that \textsc{EpiCaR} provides a more robust initialization for model merging compared to standard self-training objectives, effectively preserving the base model's general knowledge while injecting calibrated reasoning capabilities.

\section{Extended Reliability Analysis}
\label{sec:appendix_reliability}

In this section, we provide the full set of reliability diagrams across all evaluated benchmarks: MATH \citep{hendrycks2021measuring}, GSM8K \citep{cobbe2021training}, and MBPP \citep{austin2021program}. These diagrams visualize the relationship between the model's verbalized confidence and its empirical accuracy, providing qualitative evidence for the calibration improvements discussed in the main text.

We partition the model's predictions into $M=10$ equally-spaced bins based on verbalized confidence. A perfectly calibrated model would align with the dashed diagonal line ($y=x$). 

\begin{figure*}[ht]
    \centering
    \includegraphics[width=0.8\linewidth]{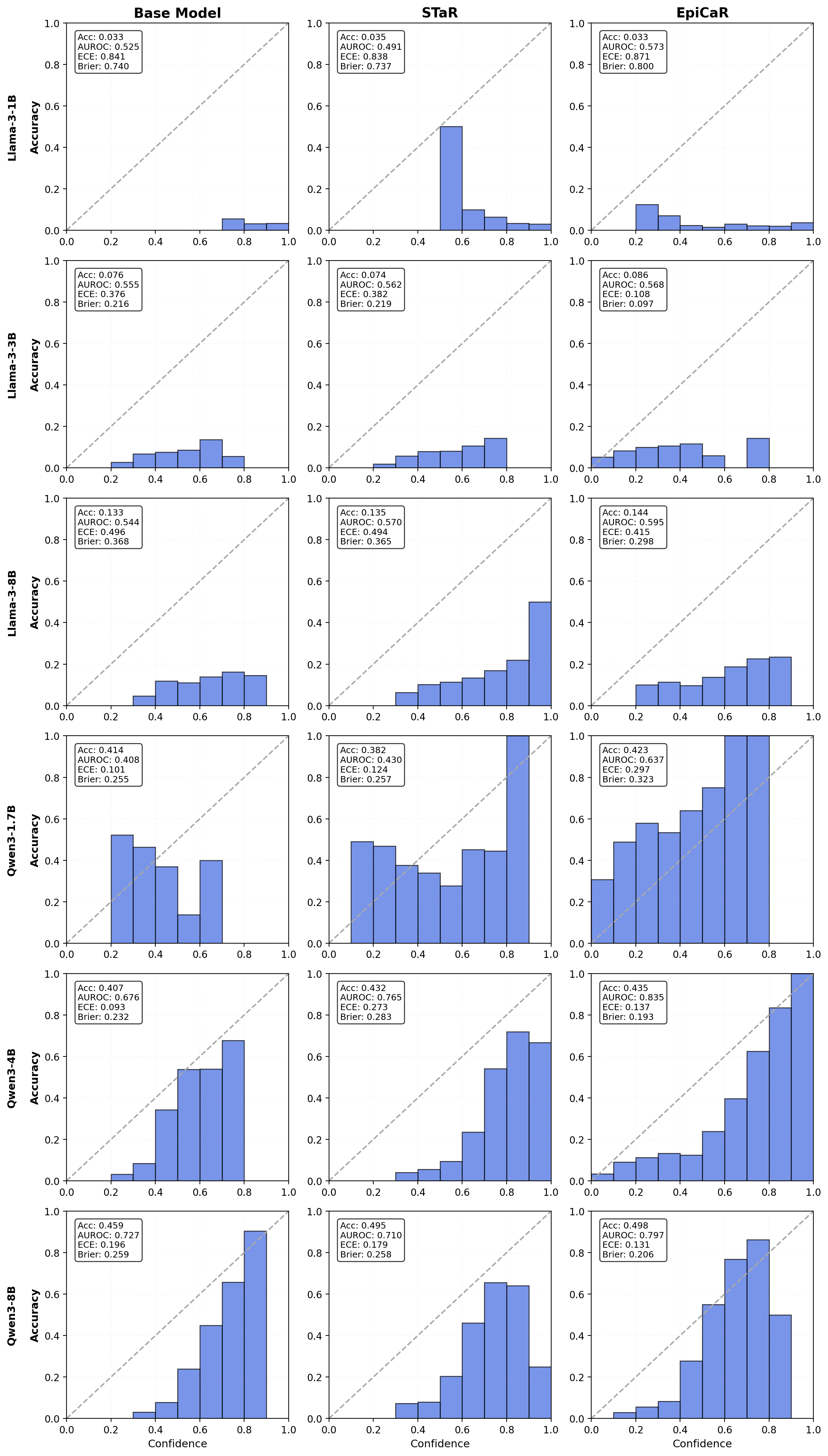}
    \caption{\textbf{Reliability Diagram: MATH (Standard).} Comparison of calibration performance between the base model, STaR, and \textsc{EpiCaR} on the MATH dataset.}
    \label{fig:rel_math_standard}
\end{figure*}

\begin{figure*}[ht]
    \centering
    \includegraphics[width=0.8\linewidth]{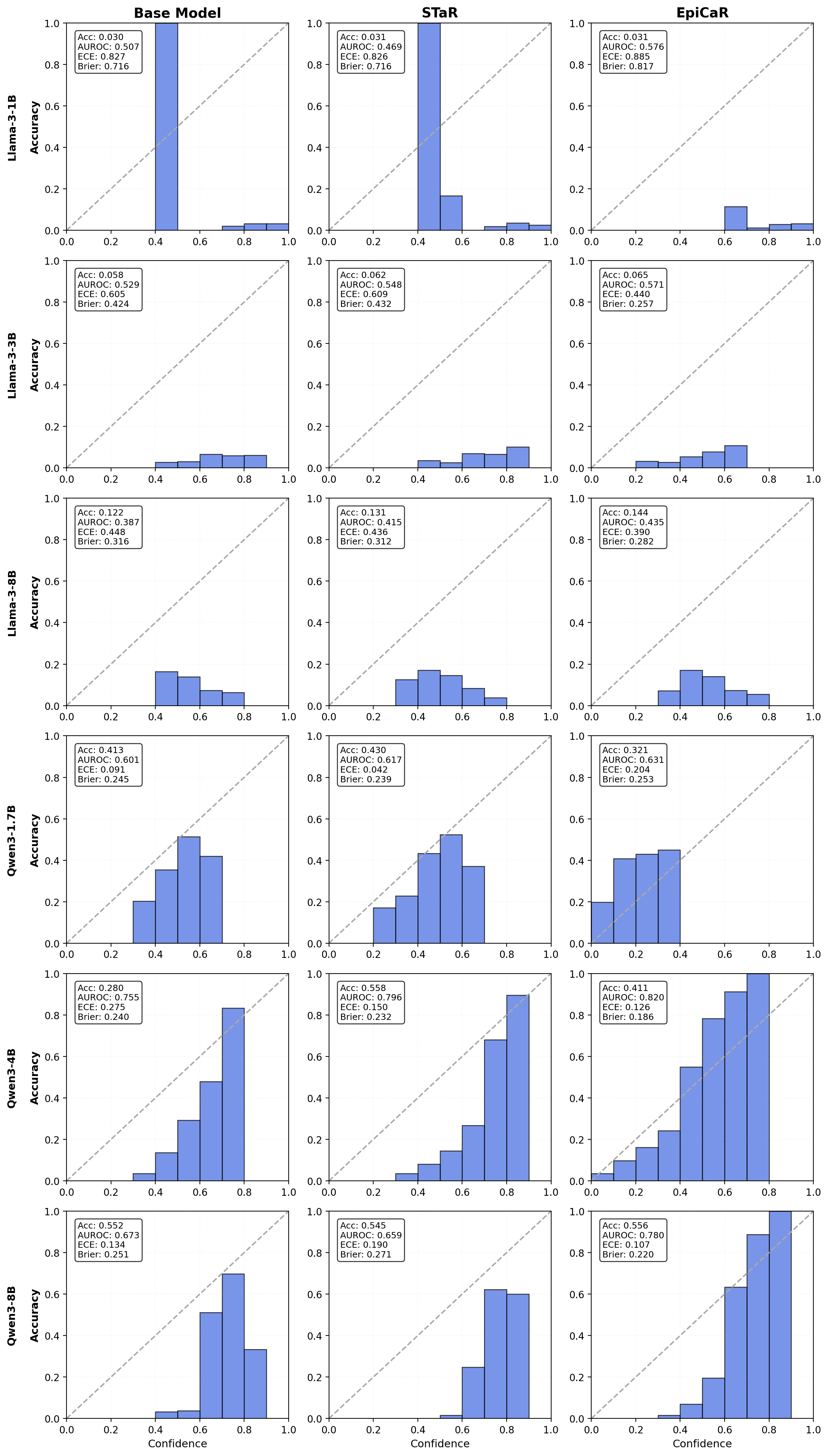}
    \caption{\textbf{Reliability Diagram: MATH (Slow Thinking).} Visualizing how internalized calibration interacts with inference-time "slow thinking" behaviors.}
    \label{fig:rel_math_slow}
\end{figure*}

\begin{figure*}[ht]
    \centering
    \includegraphics[width=0.8\linewidth]{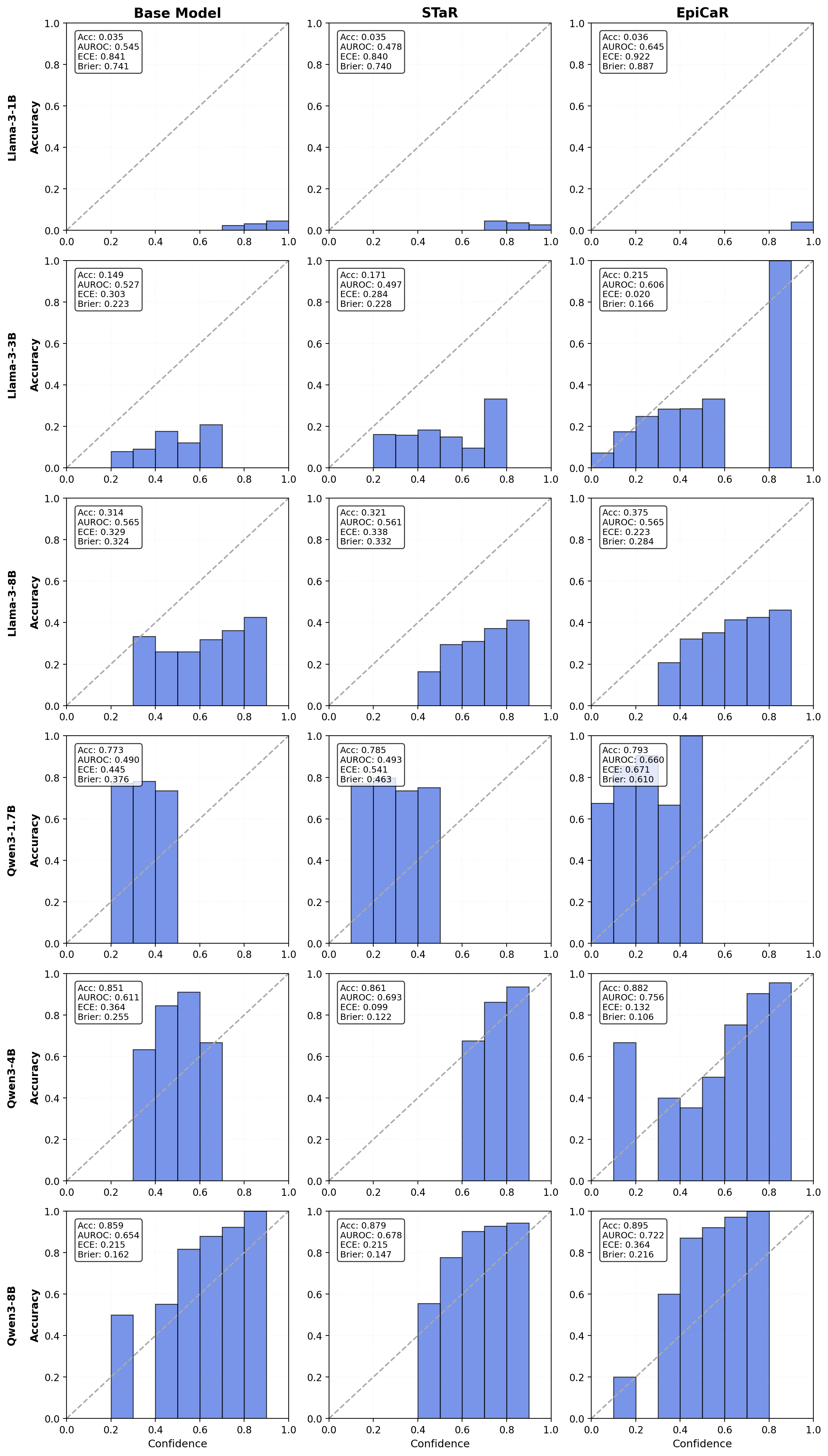}
    \caption{\textbf{Reliability Diagram: GSM8K (Zero-Shot).} Evaluation of epistemic uncertainty calibration in an out-of-distribution mathematical reasoning context.}
    \label{fig:rel_gsm8k}
\end{figure*}

\begin{figure*}[ht]
    \centering
    \includegraphics[width=0.8\linewidth]{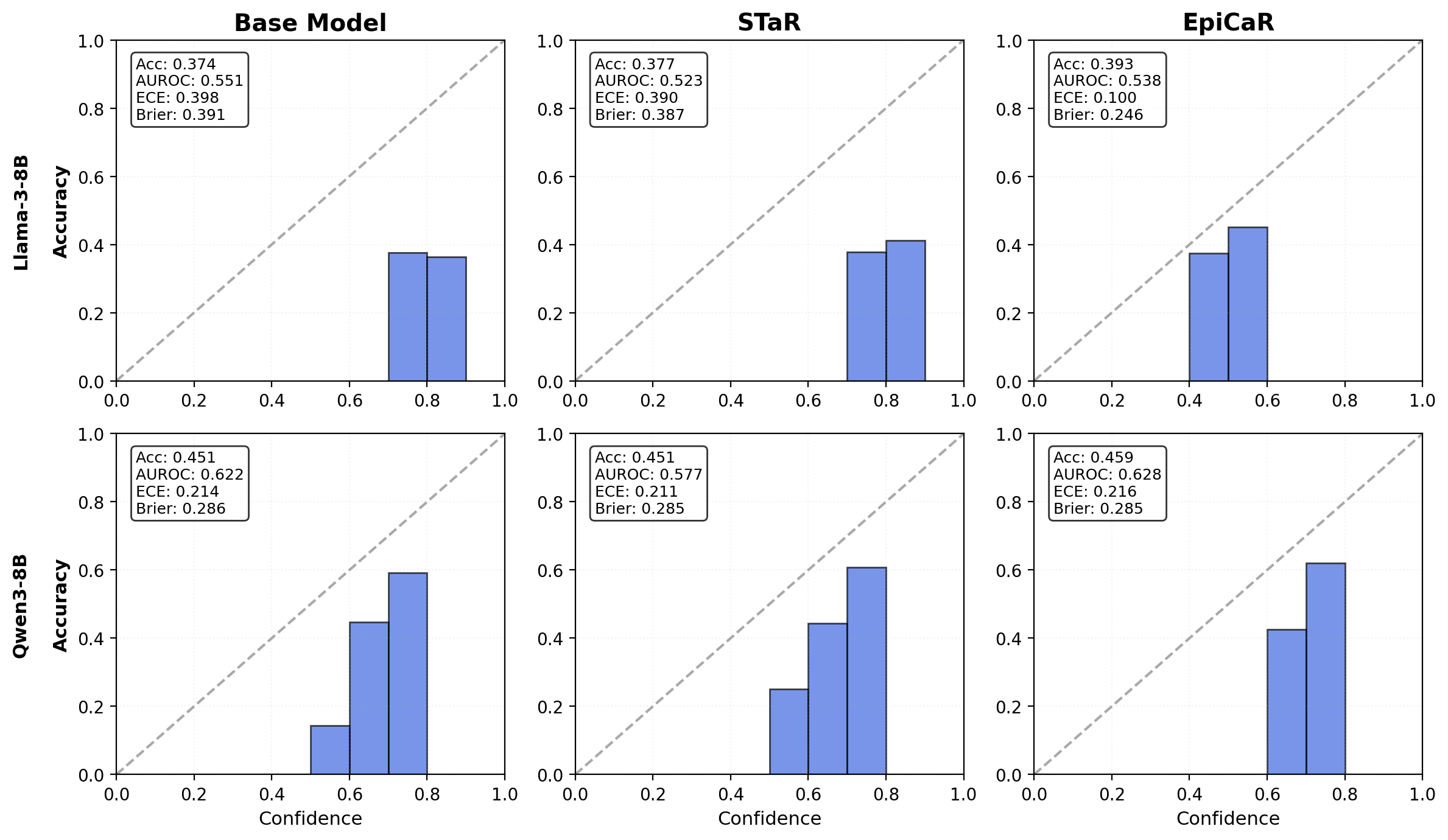}
    \caption{\textbf{Reliability Diagram: MBPP (Code Generation).} Cross-domain robustness of \textsc{EpiCaR} in programming tasks.}
    \label{fig:rel_mbpp}
\end{figure*}

As shown in Figures \ref{fig:rel_math_standard} through \ref{fig:rel_mbpp}, our method (\textsc{EpiCaR}) consistently demonstrates better alignment and higher discriminative power (AUROC) compared to the STaR baseline, which frequently exhibits overconfident clusters in the lower accuracy bins—a clear symptom of the \textit{calibration cost} of standard iterative SFT.

\begin{table*}[ht]

\centering
\small
\renewcommand{\arraystretch}{1.0}
\setlength{\tabcolsep}{10pt}
\captionsetup{width=0.9\textwidth, justification=centering}
\begin{tabular}{l l c c c c c}

\toprule
\textbf{Model} & \textbf{Method} & \boldmath$\lambda$ & \textbf{Acc. ($\uparrow$)} & \textbf{AUROC ($\uparrow$)} & \textbf{ECE ($\downarrow$)} & \textbf{Brier ($\downarrow$)} \\
\midrule\midrule
\multirow{11}{*}{Llama-3-1B} 
& Base Model & 0.0 & 3.30\% & 0.525 & 0.841 & 0.740 \\
\cmidrule{2-7}
& \multirow{5}{*}{STaR + Merging} 
& 0.2 & 3.34\% & 0.530 & 0.841 & 0.740 \\
& & 0.4 & 3.18\% & 0.529 & 0.842 & 0.741 \\
& & 0.6 & 3.54\% & 0.518 & \textbf{0.838} & \textbf{0.737} \\
& & 0.8 & 3.50\% & 0.506 & \textbf{0.838} & \textbf{0.737} \\
& & 1.0 & 3.46\% & 0.491 & \textbf{0.838} & \textbf{0.737} \\
\cmidrule{2-7}
& \multirow{5}{*}{\textbf{Ours (\textsc{EpiCaR}) + Merging}} 
& 0.2 & 3.20\% & 0.531 & 0.845 & 0.745 \\
& & 0.4 & 3.53\% & 0.555 & 0.848 & 0.754 \\
& & 0.6 & \textbf{3.62\%} & 0.551 & 0.862 & 0.779 \\
& & 0.8 & 3.44\% & \textbf{0.595} & 0.880 & 0.809 \\
& & 1.0 & 3.30\% & 0.573 & 0.871 & 0.800 \\
\midrule
\multirow{11}{*}{Llama-3-3B} 
& Base Model & 0.0 & 7.56\% & 0.555 & 0.376 & 0.216 \\
\cmidrule{2-7}
& \multirow{5}{*}{STaR + Merging} 
& 0.2 & 7.60\% & 0.562 & 0.377 & 0.217 \\
& & 0.4 & 7.46\% & 0.564 & 0.378 & 0.217 \\
& & 0.6 & 7.44\% & 0.562 & 0.378 & 0.217 \\
& & 0.8 & 7.38\% & 0.542 & 0.380 & 0.219 \\
& & 1.0 & 7.38\% & 0.562 & 0.382 & 0.219 \\
\cmidrule{2-7}
& \multirow{5}{*}{\textbf{Ours (\textsc{EpiCaR}) + Merging}} 
& 0.2 & 7.70\% & 0.568 & 0.353 & 0.200 \\
& & 0.4 & 7.82\% & 0.559 & 0.301 & 0.168 \\
& & 0.6 & 7.80\% & 0.552 & 0.237 & 0.134 \\
& & 0.8 & 7.86\% & \textbf{0.593} & 0.167 & 0.106 \\
& & 1.0 & \textbf{8.58\%} & 0.568 & \textbf{0.108} & \textbf{0.097} \\
\midrule
\multirow{11}{*}{Llama-3-8B} 
& Base Model & 0.0 & 13.30\% & 0.544 & 0.496 & 0.368 \\
\cmidrule{2-7}
& \multirow{5}{*}{STaR + Merging} 
& 0.2 & 13.26\% & 0.531 & 0.496 & 0.368 \\
& & 0.4 & 13.54\% & 0.546 & 0.495 & 0.367 \\
& & 0.6 & 13.74\% & 0.556 & 0.494 & 0.367 \\
& & 0.8 & 13.72\% & 0.555 & 0.492 & 0.365 \\
& & 1.0 & 13.46\% & 0.570 & 0.494 & 0.365 \\
\cmidrule{2-7}
& \multirow{5}{*}{\textbf{Ours (\textsc{EpiCaR}) + Merging}} 
& 0.2 & 13.90\% & 0.552 & 0.486 & 0.361 \\
& & 0.4 & 14.12\% & 0.553 & 0.470 & 0.348 \\
& & 0.6 & \textbf{15.02\%} & 0.571 & 0.443 & 0.328 \\
& & 0.8 & 14.04\% & 0.592 & 0.434 & 0.312 \\
& & 1.0 & 14.42\% & \textbf{0.595} & \textbf{0.415} & \textbf{0.298} \\
\bottomrule
\end{tabular}
\caption{\textbf{Full Results for Llama-3 Family with Model Merging.} Accuracy and calibration metrics across all $\lambda$ values for Llama-3 (1B, 3B, and 8B).}
\label{tab:llama_merging_full}
\end{table*}

\begin{table*}[ht]
\centering
\small
\renewcommand{\arraystretch}{1.0}
\setlength{\tabcolsep}{10pt}
\captionsetup{width=0.9\textwidth, justification=centering}
\begin{tabular}{l l c c c c c}
\toprule
\textbf{Model} & \textbf{Method} & \boldmath$\lambda$ & \textbf{Acc. ($\uparrow$)} & \textbf{AUROC ($\uparrow$)} & \textbf{ECE ($\downarrow$)} & \textbf{Brier ($\downarrow$)} \\
\midrule\midrule
\multirow{11}{*}{Qwen-3-1.7B} 
& Base Model & 0.0 & 41.44\% & 0.408 & 0.101 & \textbf{0.255} \\
\cmidrule{2-7}
& \multirow{5}{*}{STaR + Merging} 
& 0.2 & 41.26\% & 0.407 & 0.120 & 0.256 \\
& & 0.4 & 42.02\% & 0.413 & 0.111 & 0.257 \\
& & 0.6 & 41.64\% & 0.409 & 0.121 & 0.258 \\
& & 0.8 & 41.24\% & 0.421 & 0.123 & 0.261 \\
& & 1.0 & 38.16\% & 0.430 & 0.124 & 0.257 \\
\cmidrule{2-7}
& \multirow{5}{*}{\textbf{Ours (\textsc{EpiCaR}) + Merging}} 
& 0.2 & 42.46\% & 0.438 & 0.101 & 0.257 \\
& & 0.4 & 42.42\% & 0.497 & \textbf{0.099} & 0.257 \\
& & 0.6 & \textbf{43.08\%} & 0.543 & 0.161 & 0.270 \\
& & 0.8 & 42.72\% & 0.600 & 0.214 & 0.284 \\
& & 1.0 & 42.34\% & \textbf{0.637} & 0.297 & 0.323 \\
\midrule
\multirow{11}{*}{Qwen-3-4B} 
& Base Model & 0.0 & 40.66\% & 0.676 & \textbf{0.093} & 0.232 \\
\cmidrule{2-7}
& \multirow{5}{*}{STaR + Merging} 
& 0.2 & 41.78\% & 0.697 & 0.102 & 0.233 \\
& & 0.4 & 41.60\% & 0.746 & 0.143 & 0.235 \\
& & 0.6 & 42.32\% & 0.760 & 0.191 & 0.248 \\
& & 0.8 & 42.58\% & 0.784 & 0.240 & 0.263 \\
& & 1.0 & 43.20\% & 0.765 & 0.273 & 0.283 \\
\cmidrule{2-7}
& \multirow{5}{*}{\textbf{Ours (\textsc{EpiCaR}) + Merging}} 
& 0.2 & 42.20\% & 0.704 & 0.097 & 0.230 \\
& & 0.4 & 42.72\% & 0.782 & 0.138 & 0.220 \\
& & 0.6 & 43.40\% & 0.804 & 0.170 & 0.222 \\
& & 0.8 & \textbf{43.54\%} & 0.826 & 0.176 & 0.211 \\
& & 1.0 & 43.50\% & \textbf{0.835} & 0.137 & \textbf{0.193} \\
\midrule
\multirow{11}{*}{Qwen-3-8B} 
& Base Model & 0.0 & 45.86\% & 0.727 & 0.196 & 0.259 \\
\cmidrule{2-7}
& \multirow{5}{*}{STaR + Merging} 
& 0.2 & 45.84\% & 0.714 & 0.200 & 0.264 \\
& & 0.4 & 47.14\% & 0.721 & 0.193 & 0.261 \\
& & 0.6 & 47.44\% & 0.727 & 0.197 & 0.262 \\
& & 0.8 & 48.08\% & 0.712 & 0.190 & 0.261 \\
& & 1.0 & 49.52\% & 0.710 & 0.179 & 0.258 \\
\cmidrule{2-7}
& \multirow{5}{*}{\textbf{Ours (\textsc{EpiCaR}) + Merging}} 
& 0.2 & 46.48\% & 0.719 & 0.175 & 0.254 \\
& & 0.4 & 47.10\% & 0.729 & 0.145 & 0.243 \\
& & 0.6 & 48.54\% & 0.746 & \textbf{0.112} & 0.229 \\
& & 0.8 & 49.72\% & 0.769 & 0.123 & 0.217 \\
& & 1.0 & \textbf{49.76\%} & \textbf{0.797} & 0.131 & \textbf{0.206} \\
\bottomrule
\end{tabular}
\caption{\textbf{Full Results for Qwen-3 Family with Model Merging.} Accuracy and calibration metrics across all $\lambda$ values for Qwen-3 (1.7B, 4B, and 8B).}
\label{tab:qwen_merging_full}
\end{table*}

\clearpage

\end{document}